\documentclass[twoside,11pt]{article}
\usepackage[preprint]{jmlr2e}

 \renewcommand{\cite}{\citep}	

\usepackage{hyperref}
\usepackage{xcolor}
\usepackage{booktabs}
\usepackage{tabularx}
\usepackage{makecell}
\usepackage{amsmath,amssymb}
\usepackage{soul}
\usepackage{tcolorbox}
\usepackage{arydshln}

\definecolor{darkblue}{rgb}{0, 0, 0.5}
\hypersetup{colorlinks=true,citecolor=darkblue, linkcolor=darkblue, urlcolor=darkblue}

\newcommand{\limit}[1]{\begin{tcolorbox} #1\end{tcolorbox}}

\newcommand{\se}[1]{\textcolor{black}{#1}}
\newcommand{\sen}[1]{\textcolor{black}{#1}}
\newcommand{\cl}[1]{\textcolor{black}{#1}}
\newcommand{\clii}[1]{\textcolor{black}{#1}}
\newcommand{\cliii}[1]{\textcolor{black}{#1}}
\newcommand{\revision}[1]{\textcolor{black}{#1}}

\newcommand{\green}[1]{\textcolor{green}{#1}}
\newcommand{\red}[1]{\textcolor{red}{#1}}

\usepackage{tikz, relsize}
\definecolor{myCyan} {HTML}{ADEFFF}
\usetikzlibrary{shapes.geometric, shapes.multipart, arrows, arrows.meta, positioning, calc, chains, backgrounds}

\usepackage{stmaryrd}
\usepackage{svg}

\usepackage{lastpage}

\hypersetup{hidelinks}

\jmlrheading{25}{2024}{1-\pageref{LastPage}}{4/22; Revised 1/24}{3/24}{22-0416}{Christoph Leiter, Piyawat Lertvittayakumjorn, Marina Fomicheva, Wei Zhao, Yang Gao and Steffen Eger}
\ShortHeadings{Towards Explainable Evaluation Metrics For MT}{Leiter, Lertvittayakumjorn, Fomicheva, Zhao, Gao and Eger}
\begin{document}

\title{Towards Explainable Evaluation Metrics for Machine Translation}

\author{\name Christoph Leiter
        \email christoph.leiter@uni-mannheim.de\\
        \addr Natural Language Learning Group\\University of Mannheim\\B6 26, 68159 Mannheim, Germany \AND
        \name Piyawat Lertvittayakumjorn
        \email pl1515@imperial.ac.uk \\
        \addr Imperial College London \AND
        \name Marina Fomicheva
        \email m.fomicheva@sheffield.ac.uk\\
        \addr University of Sheffield \AND
        \name Wei Zhao
        \email wei.zhao@abdn.ac.uk\\
        \addr University of Aberdeen\\  Heidelberg Institute for Theoretical Studies \AND
        \name Yang Gao
        \email gaostayyang@google.com\\
        \addr Royal Holloway, University of London \AND
        \name Steffen Eger
        \email steffen.eger@uni-mannheim.de\\
        \addr University of Mannheim
        }

\firstpageno{1}

\editor{Ivan Titov}

\maketitle
\begin{abstract}%
Unlike classical lexical overlap metrics such as BLEU, most current evaluation metrics for machine translation (for example, COMET or BERTScore) are based on black-box large language models. They often achieve strong correlations with human judgments, but recent research indicates that the lower-quality classical metrics remain dominant, one of the potential reasons being that their decision processes are more transparent. To foster more widespread acceptance of novel high-quality metrics, explainability thus becomes crucial. In this concept paper, we identify key properties as well as key goals of \textit{explainable machine translation metrics} and provide a comprehensive synthesis of recent techniques, relating them to our established goals and properties. In this context, we also discuss the latest state-of-the-art approaches to explainable metrics based on generative models such as ChatGPT and GPT4. Finally, we contribute a vision of next-generation approaches, including natural language explanations. We hope that our work can help catalyze and guide future research on explainable evaluation metrics and, mediately, also contribute to better and more transparent machine translation systems. 
\end{abstract}


\section{Introduction} \label{sec:intro}
The field of evaluation metrics for Natural Language Generation (NLG), especially machine translation (MT) is \se{in a crisis} \citep{marie-etal-2021-scientific}. 
Despite the development of multiple high-quality evaluation metrics in recent years \citep[e.g.,][]{zhao-2019,zhang-2020,rei-2020,sellam-2020,yuan-etal-2021-advances,rei-etal-2023-scaling,kocmi-federmann-2023-gemba}, the Natural Language Processing (NLP) community appears hesitant to adopt them for assessing NLG systems \citep{marie-etal-2021-scientific,gehrmann2022repairing}. Empirical investigations of \citet{marie-etal-2021-scientific} indicate that the majority of MT papers relies on surface-level evaluation metrics such as BLEU and \clii{METEOR} \citep{papineni-2002, banerjee-2005}, which were created two decades ago, a trend that may allegedly have worsened in recent times.
These surface-level metrics cannot \se{(even)} measure semantic similarity of their inputs and are thus fundamentally flawed, particularly when it comes to assessing the quality of recent state-of-the-art MT systems \citep[e.g.,][]{peyrard-2019-studying, freitag-etal-2022-results}, raising concerns about the credibility of the scientific field.
We argue that the potential reasons for this neglect of recent high-quality metrics include:
(i) non-enforcement by reviewers; (ii) easier comparison to previous research, for example, by copying BLEU-based results from tables of related work  (potentially a pitfall in itself); (iii) 
computational inefficiency to run expensive new metrics at large scale; 
(iv) lack of trust in and transparency of high-quality black box metrics. In this work, we concern ourselves with the last-named reason, 
and address the \emph{explainability} of such metrics. 

In recent years, explainability has become a crucial research area in AI due to its potential benefits for users, designers, and developers of AI systems \cite{Samek2018Overview,vaughan2020human}.\footnote{\se{As of now, there is no universally accepted definition of explainability in the AI community.}
In this work, we \cliii{adapt} the definition by \citet{barredo-arrieta-2020}, which we discuss in \S\ref{sec:explainability}.}
For \emph{users} of the AI systems, explanations help them make more informed decisions
(especially in high-stake domains) 
\cite{sachan2020explainable,lertvittayakumjorn-etal-2021-supporting}, 
better understand and hence gain trust of the AI systems
\cite{pu2006trust,toreini2020relationship},
and even learn from the AI systems to accomplish 
tasks more successfully
\cite{mac2018teaching,lai2020chicago}.
For AI system \emph{designers and developers}, explanations allow them to
identify the problems and weaknesses of the system\se{s}
\cite{krause2016interacting,han-etal-2020-explaining},
calibrate the confidence of the system\se{s} \cite{zhang2020effect},
and improve \se{them} 
accordingly \cite{kulesza2015principles,lertvittayakumjorn-2021}.

Explainability is particularly desirable for evaluation metrics.
\citet{sai-2020} suggest that explainable NLG metrics should focus on providing more information than 
just a single score (such as fluency or adequacy). 
\citet{celikyilmaz-2020} stress the need for explainable evaluation metrics 
to spot system quality issues and to achieve a higher trust in the evaluation of NLG systems. 
Explanations indeed play a vital role in building trust for new evaluation metrics.\footnote{\cliii{As an illustrating example, \citet{moosavi2021scigen} distrust using the metric BERTScore \cite{zhang-2020} applied to a novel text generation task, as it assigns a score of 0.74 to a nonsensical output, which could be taken as an unreasonably high value. While BERTScore may indeed be unsuitable for their novel task, the score of 0.74 is meaningless here, as evaluation metrics may have arbitrary ranges.} In fact, BERTScore typically has a particularly narrow range, so a score of 0.74 even for bad outputs may not be surprising \se{(BERTScore provides a scaling method that solves these issues to some degree)}. 
Explainability techniques would be very helpful in preventing such misunderstandings, \cl{e.g., by showing that the features BERTScore attends to align with human judgement}. }
For instance, if explanations of the scores align closely with human reasoning and faithfully reflect a metric's internal decision process, this metric may be more likely to gain acceptance within the research community (see \S\ref{sec:explainability} for a definition of faithfulness and related terminology).

By contrast, if \cl{faithfulness is given and} the explanations are counter-intuitive, users and developers \cl{might} lower their trust and be alerted to take additional actions, such as trying to improve the metrics using insights from the explanations or looking for alternative metrics that are more trustworthy. 
Furthermore, explainable metrics can be used for other purposes:
for example, when a metric produces a low score for a given input, highlighted words (a widely used method for explanation, see \S\ref{sec:taxonomy}) in the input are natural candidates for manual post-editing \revision{(when working with reference-free metrics, see \S\ref{sec:metrics})}. 

This concept paper aims at providing a systematic overview of the existing efforts
in explainable MT evaluation metrics and an outlook for promising future research directions.
The main focus of this work lies on explainable evaluation metrics for MT, although many of our observations and discussions can \se{likely} be adapted to other NLG tasks. 

We make the following contributions, outlined by the structure of our paper:\\

\textbf{\S\ref{sec:terminology} Background \& Terminology:} We give an overview of concepts \clii{important} \cl{for} explainable MT evaluation. We also highlight different goals and the audiences that require them. 

\textbf{\S\ref{sec:taxonomy} Taxonomy:} We provide a structured overview of previous efforts in explainable MT evaluation. \cliii{An overview} is shown in Figure \ref{fig:taxonomy}. \clii{The process of selecting works for this taxonomy is described in \S\ref{selection} \textbf{Literature Review \& Selection}.}

\textbf{\S\ref{sec:future} Future Work:} An analysis of \clii{underexplored} research \cl{directions}, \clii{and a collection of exemplary methods from other NLP domains to illustrate potential future paths. Here, we also discuss the usage of recent large language models (LLMs), like ChatGPT \cite{chatgpt} and GPT4 \cite{gpt4}.} 

\clii{We provide an overview of \cliii{\textbf{Related Work}} in \S\ref{sec:relatedwork} and a \cliii{\textbf{Conclusion}} in \S\ref{sec:conclusion}}.\\

With this work, we aim to \cl{solidify the field of explainable MT metrics} and provide guidance for researchers, e.g., metric developers, who aim to better understand and explain \cl{MT metrics}, as well as \cl{users} that want to employ explainability techniques to explain \cl{metric outputs. In \se{the} long term, we envision that our work will aid the development of improved MT metrics and thereby help to improve the quality of machine translations. Potentially, explanations can also aid other use cases such as \cliii{translation selection and semi-automatic labeling} (see \S\ref{goals_sec}).} \se{We also hope that our work will more generally inspire explainable NLG metric design.}

\section{\cl{Background \& }Terminology}
\label{sec:terminology}
In this section, we first introduce \clii{and relate} definitions and dimensions of MT metrics and explainability, which we use in the later parts of the paper. Further, we collect goals and target audiences \cliii{of explainable MT metrics}.

\subsection{Machine Translation Evaluation Metrics}\label{sec:metrics}
MT metrics grade machine translated content, the \textit{hypothesis}, based on ground truth data. \cliii{Metrics can be categorized along several dimensions \citep[e.g.,][]{sai-2020, celikyilmaz-2020}. Here, we focus on a minimal \se{specification} that we will leverage later. A summary is shown in Table \ref{table:metrics}. Some parts of this section refer to explainability (see definitions in \S\ref{sec:explainability}).}

\begin{table}[!htb]
\centering
    \begin{tabularx}{\textwidth}{lXX}
     \toprule
     \textbf{Dimension} &  \textbf{Description} \\ 
     \midrule
     \emph{Input type} & 
     Whether source, reference translation or both are used as ground truth for comparison \\ 
     \hline
     \emph{Granularity} & At which level a metric operates: Word-level, sentence-level, document-level \\ \hline
     \emph{Quality aspect} & What a metric measures: Adequacy, fluency, etc. \\ \hline
     \emph{Learning objective} & How a metric is induced: Regression, ranking, etc. \\
     \bottomrule
\end{tabularx}
\caption{A typology of categorizations for evaluation metrics.}
\label{table:metrics}
\end{table}

\noindent\textit{\\Input type} \;\;\;\;
We call a metric \textit{reference-based} if it requires one or multiple human reference translations to compare with the hypothesis as ground truth, which can be seen as a form of supervision \cl{\citep[e.g.,][]{papineni-2002,zhao-2019,rei-etal-2022-comet}}.  
\textit{Reference-free} metrics do not require a reference translation to grade the hypothesis \citep[e.g.,][]{zhao-2020,ranasinghe-2020,belouadi2023uscore}. 
Instead, they directly compare the source to the hypothesis. 
In the literature, reference-free MT evaluation is \se{sometimes} also referred to as ``reference-less'' \citep[e.g.,][]{mathur-2020} or 
``quality estimation'' \citep[e.g.,][]{zerva-EtAl-2022-WMT}.  
This dimension is important for explainable MT evaluation, as explainability techniques might need to consider which resources are available.

\noindent\textit{\\Granularity} \;\;\;\;
Translation quality can be evaluated at different levels of granularity: word-level, sentence-level and document-level.
The majority of metrics for MT return a single sentence-level score for each input sentence \cl{\citep[e.g.,][]{zhao-2019,zhang-2020}}. Beyond individual sentences, a metric may also score whole documents (multiple sentences) \citep{jiang-etal-2022-blonde,zhao:2022-discoscore}. 
In the MT community, metrics that evaluate translations at the word-level (for example, whether individual words are correct or not) are also common \citep{Turchi2014AdaptiveQE,shenoy2021investigating}. \cliii{Recently, there has been a tendency to compare metrics
to human scores derived from fine-grained error analysis as they tend to correspond better to human professional translators \citep{freitag-etal-2021-experts}.} \cl{Metrics of higher granularity (e.g., word-level) provide more explanatory value than those of lower granularity, as they provide more details on which parts of the input might be translated incorrectly \cl{\citep[e.g.,][]{leiter-2021-reference,fomicheva-2021,fomicheva-etal-2022-translation}}.} Due to their prevalence, most techniques we describe in \S\ref{sec:taxonomy} target the explainability of \revision{metrics evaluated on sentence-level.}

\noindent\textit{\\Quality aspect} \;\;\;\; This refers to the properties a metric evaluates and measures, for example, human assigned adequacy (e.g., via direct assessment, see \citet{graham-2016}, obtained from crowd-workers), fluency, or other aspects (such as relevance, informativeness, or correctness, mostly in other NLG fields) \cl{\citep[e.g.,][]{sai-2020, celikyilmaz-2020}}. 
We use this dimension in \cliii{\S\ref{sec:taxonomy}}, as metrics that report multiple quality aspects provide more information on specific translations, i.e., they have more explanatory value for potential users. 

\noindent\textit{\\Learning objective} \;\;\;\;
\citet{yuan-etal-2021-advances} describe a further distinction based on the task a metric is designed to solve (or \emph{how} a metric is implemented) to achieve high correlation with human judgments. They identify the three tasks below.\footnote{\se{\citet{chen2022menli} argue for metrics induced from natural language inference (NLI) as  a conceptual framework.}} 
\begin{itemize}
    \item \textbf{Unsupervised Matching:} They place all metrics that match hypothesis and reference tokens in this dimension, e.g., BLEU \citep{papineni-2002} and BERTScore \citep{zhang-2020}.
    \item \textbf{Supervised Regression/Ranking:} These models use supervised regression to train a metric to predict continuous values that directly correlate with human judgments from hypothesis and ground truth tokens.
    Alternatively, a metric may be trained to rank hypothesis sentences. For example, \citet{rei-2020} propose a ranking setup in which they minimize a loss that becomes smaller the greater the distance between embedding representations of the two sentences become. \se{Especially in the MT community, metrics trained on human scores (e.g., COMET ;\citealt{rei-etal-2022-comet}) have become more and more dominant, while \citet{belouadi2023uscore} argue that fully unsupervised metrics\revision{, i.e., metrics that were constructed without reference sentences, parallel data and human scores,} are more widely applicable.} 
    \item \textbf{Text Generation:} 
    Some metrics use the probability that a sentence is created by a text generation system \revision{with a paraphrasing or translation task} as sentence-level score. \revision{For example, the probability of translating a source sentence into a hypothesis sentence with an MT model can be used as score.}
    \citet{yuan-etal-2021-advances}
    place the metric PRISM \citep{thompson-2020} and their own metric BARTScore in this \revision{category.}  In contrast, some recent metrics are based on in-context learning and directly generate scores \citep[e.g.][]{kocmi-federmann-2023-gemba, lu2023error}.
\end{itemize}
\cl{The distinction \cl{of learning objectives} is important for explainable MT evaluation, as the applicability of some explainers depends on the objective. 
}

\bigskip

\begin{figure}[htb]
    \centering
    \includegraphics[width=\textwidth]{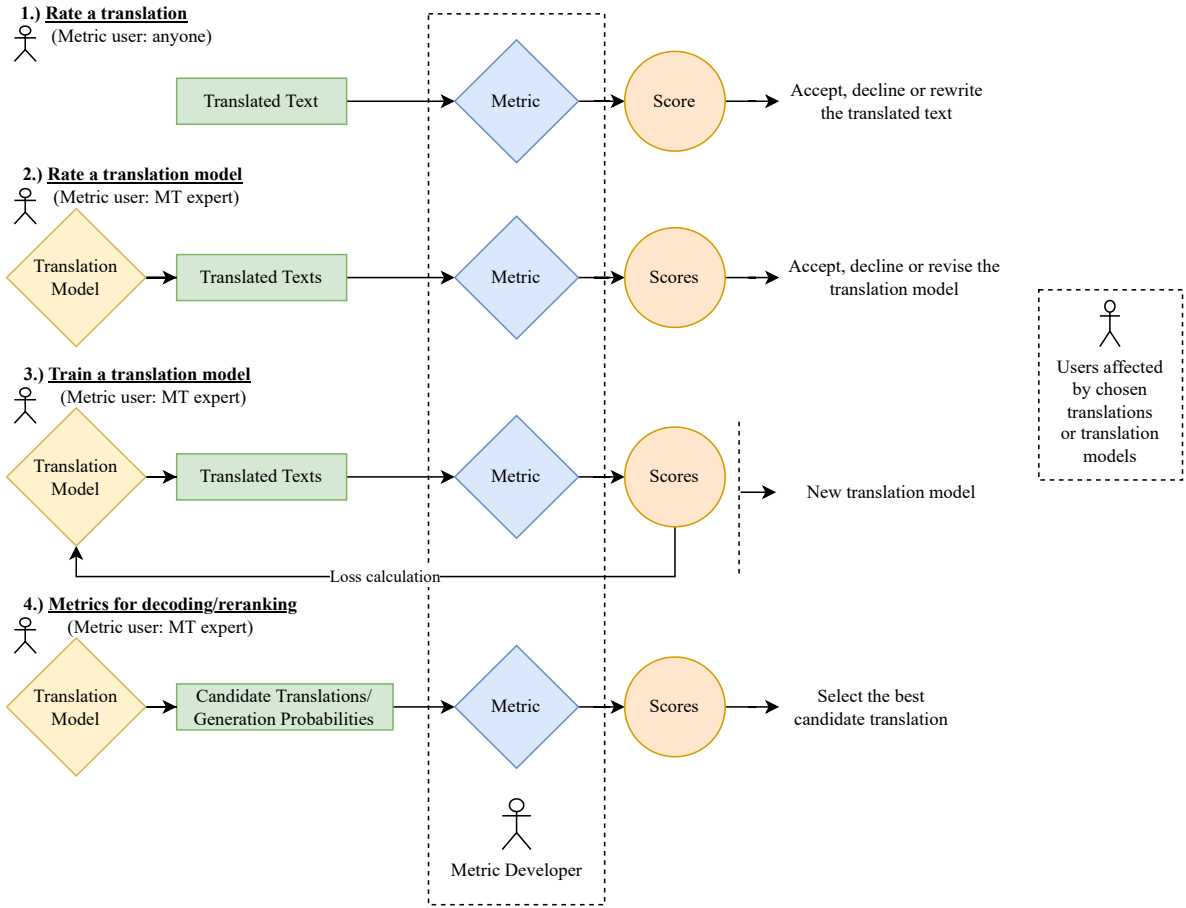}
    \caption{An overview of use-cases for MT metrics. In number 4, many candidate translations are produced for a single source. One of them is chosen as \cliii{the} main translation, \se{via the metric}.} 
    \label{fig:MTUseCases}
\end{figure}

\revision{Sentence-level} MT metrics \clii{also} have multiple use cases;  
see Figure \ref{fig:MTUseCases} \cl{(in the following, numbers \revision{written in brackets} correspond to this figure)}. \cl{(1)} One use case is to rate a single translated text, for example, in a scenario where a human user wants to check their own human\footnote{\cliii{Note, however, that humans might make different types of translation errors \citep{Specia2018}. Therefore, metric performance first needs to be evaluated for this use case.}} translation or a translation they received by an \se{MT} model before using it in a downstream task \citep[e.g.,][]{murgolo-etal-2022-quality,zouhar-etal-2023-poor}. This setup can also be used to filter parallel corpora for high-quality training data of new MT models \citep[e.g.,][]{ramos2023aligning,peter-etal-2023-theres}. \cl{(2)} \revision{A second, common} use case is to rate a large number of translations \revision{of different texts by one MT system to obtain a system-level score (for example by averaging sentence-level scores)} \cite[e.g.,][]{freitag-etal-2022-results}. \revision{Then, different MT systems can be compared their system-level scores that were computed on the same corpus}. \cl{(3)} Another use case is to employ metrics in the training process of MT models, e.g., using reinforcement learning \cite[e.g.,][]{wu-etal-2018-study} or optimizing by computing derivatives with Gumbel-softmax \cite[e.g.,][]{jauregi-unanue-etal-2021-berttune}. 
\cl{(4)} \cliii{The} fourth use case is to use the metrics in the decoding process \se{of} \cliii{machine translation systems} for reranking or with minimum Bayes risk decoding \cite{fernandes-etal-2022-quality}. 

The use cases can have different audiences and explainability might play a different role to them (see \ref{sec:audience}). For example, anyone might be a user that wants to rate a translation, but usually only \revision{certain user groups, such as translation experts or MT experts,} want to rate a translation model. 

\subsection{Explainability}
\label{sec:explainability}
In this section, we \cliii{(1)} define explainability in the scope of this work, (2) discuss the definitions and roles of faithfulness and plausibility for explainable MT metrics and \cliii{(3)} contrast \cliii{explainability} with the terms robustness and fairness. Then, we \cliii{(4)} discuss the properties we use to structure and describe the taxonomy \clii{of prior works on explainable MT metrics} (see \S\ref{sec:taxonomy}), namely \textit{explanation level}, \textit{explanation type}, \textit{model access} and \textit{audiences}.  
\cliii{Based on (1), we decide on the scope of papers we include in the taxonomy in \S\ref{selection}.}

\noindent\textit{\\Explainability \& interpretability} \;\;\;\;
Prior work has noted that the terms \textit{interpretability} and \textit{explainability} vary in definition and are sometimes used ambiguously throughout the literature \citep{barredo-arrieta-2020, jacovi-2020, VILONE202189, freiesleben2023}. In this paper, we follow the notion of \emph{passive interpretability} and \emph{active explainability} \citep[for example,][]{barredo-arrieta-2020}.  
In particular, \cliii{we adapt the definition of \citet{barredo-arrieta-2020} as follows:} 
\begin{quote}
An \cliii{\textit{MT metric}} is explainable if \cliii{itself} or some external instance, an \textit{explainer}, can actively provide explanations that make \cliii{its} decision process \cliii{or its output} clearer to a certain audience. 
\end{quote}
\cliii{This definition varies from \citet{barredo-arrieta-2020} in two aspects: (1) It includes \textit{external explainers} (also called post-hoc methods) and (2) it includes explanations of the metric's output. We note that (2) is different from explaining the metric's decision process, as output explanations may not provide insights on how the metric works. Still, we include (2) in our explainability definition for two reasons. One is that there have already been multiple shared tasks and other works on metric explainability that only evaluate how plausible explanations explain the output of a metric \citep[e.g.,][]{fomicheva-2020b, fomicheva-2021, zerva-EtAl-2022-WMT, xu2023instructscore}.   
We want to accommodate these works in our definition. The other reason is that there are use cases for such explanations of the output (specifically for the goals of making metrics more accessible and semi-automatic labeling; see \S\ref{goals_sec}).
}

In contrast, a \cliii{metric} is interpretable if it can be passively understood by humans (for example, decision trees and k-nearest neighbors) \citep{barredo-arrieta-2020}. Modern MT metrics are mostly based on embeddings and black box language models, making the\se{m} 
non-interpretable.
The papers we categorize in this work generate explanations for MT metrics \cl{and their outputs}, i.e., they consider explainability.

\noindent\textit{\\Faithfulness \& plausibility} \;\;\;\;
\revision{An explanation is \textit{faithful} if it accurately represents the reasoning process underlying a model’s prediction(s) \citep{jacovi-2020}. In contrast, an explanation is \textit{plausible} if it convinces its audience that a model's prediction is correct \citep{jacovi-goldberg-2021-aligning}, i.e., if it aligns with human judgment. We illustrate the properties with the following example:}
\begin{itemize}
    \item \textbf{Source}: Things became complicated
    \item \textbf{Hypothesis}: Die Dinge bekamen kompliziert
    \item \textbf{Metric(Source, Hypothesis)} = 0.8
    \item \textbf{Explanation 1}: The word \textit{became} is wrongly translated into \textit{bekamen} instead of \textit{wurden}. However, the meaning of the translation can easily be inferred. Therefore, on a scale of 0 to 1, a score of 0.8 is given.
    \item \textbf{Explanation 2}: The words \textit{Die Dinge} have 8 letters. Therefore,  a score of 0.8 is assigned. 
\end{itemize}

\revision{In this example, the English source sentence was (wrongly) translated into the German hypothesis sentence. A metric grades the translation with a score of 0.8 and two explanations (here, natural language explanations) are provided. Comparing these explanations, explanation 1 is a more plausible description (to translation experts) of why the score of 0.8 was assigned, as it bases the score on translation errors. However, which of the explanations describes the metric more faithfully cannot be inferred from the text. For example, internally the metric could be relying on word lengths, making explanation 2 more faithful, even though it is linguistically non-sensical.} 

\revision{When explainability techniques are used for metrics, faithfulness is always desirable and often, depending on the goal of the explanations, required. Plausibility (to translation experts), on the other hand, is required for some goals and not important for others (see \S\ref{goals_sec}).}

\revision{As a side note, ``complete faithfulness'' of an explainability technique may never be established because it would require capturing the full decision process, e.g., of neural networks, in human understandable descriptions. For example, \citet{jacovi-2020} deem it as ``a \textit{unicorn}
which will likely never be found''.}

\noindent\textit{\\Fairness \& robustness} \;\;\;\; \revision{We also categorize methods that \textbf{test} the \textit{robustness} and \textit{fairness} of a model (here, a metric) as explainability techniques. This is because these tests produce explanations that will help certain audiences to understand a model's behavior better \cite{barredo-arrieta-2020}. 
\\Here, robustness measures how well a model can address execution errors, erroneous and malicious inputs and unseen data \citep[e.g.,][]{10.1145/3555803}.
In contrast, \citet{10.1145/3555803} define fairness as the goal that a model should not have biases that result in unfair treatment for certain groups of people.\footnote{There are many related and partly conflicting definitions and terms in explainable AI research \cite[e.g.,][]{VILONE202189} and the given definitions are not yet firmly established in the literature. For example, \citet{Sharma2020CERTIFAIAC} define explainability, robustness and fairness as exclusive properties. The main focus of our work is to build a taxonomy of methods for explainable MT metrics and the chosen definitions allow us to reasonably relate existing work.}} 

\revision{In the MT metrics literature, this overlap in terminology is apparent for techniques that check how robust metrics are to input perturbations. Here, some works frame these as explainability techniques \cite[e.g.,][]{sai-2021, karpinska-etal-2022-demetr}, while others frame them as robustness tests \cite[e.g.,][]{he-etal-2022-on, chen2022menli}; see \S\ref{sec:taxonomy}.}

\revision{Robustness could be measured coarse-grained\se{ly} with a single score, such as the minimal number of adversarial perturbations required to successfully attack a \cliii{metric}, or fine-grained\se{ly}, by using predefined perturbation templates \cite[e.g.,][]{sai-2021} or attackers that will attack with perturbations of specific properties \cite[e.g.,][]{chen2022menli}. While a single score can also explain whether \cliii{metric} behavior is robust, susceptibility scores to various properties allow to assess \cliii{metrics} in greater detail. The same is true for fairness \cl{\citep{sun-etal-2022-bertscore}}.}

\noindent\textit{\\Explanation level} \;\;\;\;
\label{local_global}
\cl{
Explainability 
techniques can be distinguished into \emph{decision-\\understanding} and \emph{model-understanding} \cite[e.g.,][]{Gehrmann2020,HOLZINGER2022263}.}
\cl{Decision-understanding} methods \se{are described as}  
explain\se{ing} the decision process for a specific model output with respect to a specific input \se{while} \cl{model-understanding} methods \se{are described as}  
explain\se{ing} general properties of a model that steer its decisions.
\revision{For example, a decision-understanding method could show that a metric assigned a score of 0.5, because word \textit{xyz} was translated incorrectly. In contrast, a model-understanding method could show that a metric will always return a score of 1, given certain conditions.}
\cl{Often, decision-understanding is also referred to as \emph{local} and model-understanding as \emph{global} \cite[e.g.,][]{lipton-2016, guidotti-2018,doshi-velez-2017, danilevsky-2020}.}

\noindent\textit{\\Explanation type} \;\;\;\;
Further, we distinguish explanation types, i.e., the output format of the explanations \citep[e.g.,][]{wiegreffe-2021}. One example are natural language explanations that describe the model output or its internal decision process with free text. In the case of MT metrics, this could be a sentence like ``The metric returned a score of 0.6 for the translation, as it correctly captures all details of the source sentence but has severe grammatical errors at positions [...]''. There are many possible explanation types \citep[e.g.,][]{barredo-arrieta-2020, wiegreffe-2019}, such as, feature importance explanations or counterfactuals. For our taxonomy, we select those that occur in \se{our target} papers (see \S\ref{selection}). The definition of each explanation type is paired with the description of methods in \S\ref{sec:taxonomy}.

\noindent\textit{\\Model Access} \;\;\;\;
Most taxonomies of explainability techniques consider the required level of model access \citep[e.g.,][]{barredo-arrieta-2020,linardatos-2021,madsen-etal-2022-post}. Here, techniques that require no access (that can work with any \textit{black box model}) are called \textit{model-agnostic}. Techniques that require access to model parameters are called \textit{model-specific}, \cliii{as only specific model architectures satisfy the requirements of the method}. Due to the prevalent role of transformer architectures, many modern \se{evaluation} metrics use \textit{embeddings} and \textit{attention} mechanisms and allow for \textit{gradient} computations. This allows to apply many techniques that are specific to these architectures to \se{such}  metrics. A special type of model-specific explanations are those that are given by a model (here a metric) itself. We call these metrics ``explainable by design'', as they are designed in a way that provides more details on their explanations or decision process. 
\cl{Examples for MT metrics in this category are unsupervised matching metrics like BERTScore \cite{zhang-2020} and MoverScore \cite{zhao-2019}, as they compute token-level similarity scores that can be used as explanations of the sentence-level score \cite{leiter-2021-reference, fomicheva-2021}.} 
Finally, there are also \textit{interpretable models} which are easy to understand for humans such as Naive Bayes and decision trees \cite{freitas2014comprehensible, arya-2019}. We can normally obtain decision-understanding explanations of these models at the same time as predictions (for example, the corresponding path in the decision tree), while the models can also be considered explanations of themselves (for example, the trained decision tree itself). We do not use the dimension of \textit{interpretable models} in our taxonomy \cl{(see \S\ref{fig:taxonomy})} 
as none of the state-of-the-art metrics falls into this category. 
Model-agnostic techniques require least model access. \revision{Further,} model-specific techniques, which use the gradient, embeddings or attention weights require more access. \revision{Finally,} models that are explainable by design require access to the full model to generate explanations.\footnote{Note that besides interpretable models, none of \se{the} access levels inherently leads to faithful explanations, i.e., explanations that correctly describe the internal processes of a model. Hence, this property needs to be considered separately \citep[e.g.,][]{jacovi-2020}.}

\label{sec:audience}
\noindent\textit{\\Audiences} \;\;\;\; Prior works in explainability point out that different audiences require different types of explanations \citep[e.g.,][]{barredo-arrieta-2020, Sharma2020CERTIFAIAC}. For example, non-experts might have difficulties understanding explanations given in the domain language of experts. \cl{\citet{barredo-arrieta-2020} distinguish \textit{model users}, \textit{affected users}, \textit{developers}, \textit{regulatory entities}, and \textit{managers}}. Users of MT metrics are \revision{often}  
experts in MT that want to evaluate or improve MT \se{systems}, i.e., MT developers. \revision{Another group of users are translation experts that want to compare MT systems and rate translations for usage in daily tasks 
(for example for post-editing).} 
A third group of users might be non-experts that want to use the models to check their own translations or machine generated translations. Affected users might be the ones that receive translations that were graded by a metric or generated by a model that was supervised or tested with the metric. For example, \citet{fernandes-etal-2022-quality} employ MT metrics in the decoding process of MT models. We consider users of those MT models \se{as} affected users. \cl{Developers are the designers of MT metrics. Regulatory entities and managers have an interest in enforcing guidelines.} \cliii{We describe the relation of the audiences towards goals of explainable MT metrics in \S\ref{goals_sec} and associate audiences with explanation types in \S\ref{sec:taxonomy}.} 

\subsection{\cl{Goals} of Explainable MT Metrics}
\label{goals_sec}
\begin{figure}[htb]
    \centering
    \includegraphics[width=\textwidth]{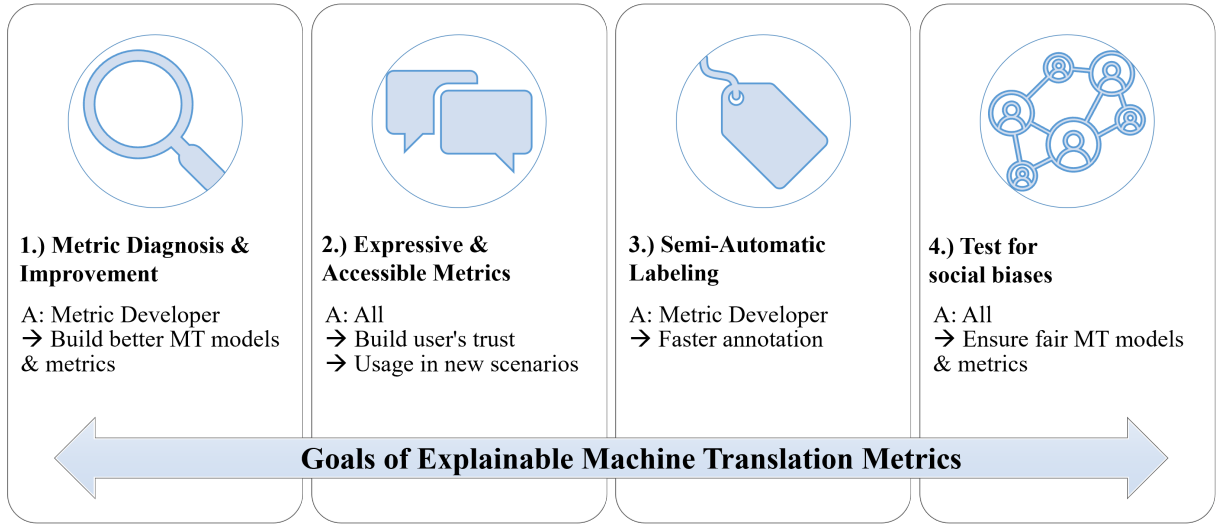}
    \caption{Goals of explainable MT evaluation. 
    Each goal has audiences, which might be interested in its fulfillment; these are indicated with ``\textit{A:}''. Further, each goal follows intentions, which are denoted with $\shortrightarrow$.}
\label{table:xMTEGoals}
\end{figure}

In this section, we discuss goals that require specific focus for explainable MT evaluation. \cliii{We orient ourselves on common goals proposed by \citet{lipton-2016}, \citet{barredo-arrieta-2020} and \citet{jacovi-goldberg-2021-aligning}, and identify four specific ones that match the literature we identify in \S\ref{selection} and use cases that might commonly occur in our field of research (see Figure \ref{table:xMTEGoals}).} \revision{Note that goal 1 and 4 address metric internals, while goal 2 and 3 mostly consider explanations of metric outputs (and with this, also the quality of the input translations).}
We follow with a short description of each of them below. 

\begin{itemize}
    \item \textbf{Goal 1: Diagnose and improve metrics.} By explaining why a metric predicted a certain score for a machine translation, where humans would assign a different score, developers might understand its weaknesses \citep[e.g.,][]{kaster-etal-2021-global, sai-2021}. This can enable architectural changes or the selection of different data sets, leading to metric improvement. Likewise, explaining the whole metric can unveil \cliii{whether it} follows desired general properties or might otherwise be led astray by carefully crafted adversarial inputs. 
    \revision{This goal requires explanations to be \textit{faithful} to some exten\se{t}, such that developers can understand the actual faults of the explained metric. \textit{Plausibility} on the other hand may be harmful and trick developers into believing that their metric is working, therefore, it should be evaluated separately \citep{jacovi-2020}.}
    \se{Goal 1} is especially important for the audience \textit{metric developers}, as they need to understand the shortcomings to improve on them.
    \item \textbf{\se{Goal 2}: Make metrics more expressive and accessible.} A metric that assigns a score on a sentence-level is difficult to understand\revision{, especially for users \revision{who} are not experienced with MT and MT metrics}. 
    If further information about a metric's output is provided, such as which words it considers incorrect, the metric may become more accessible \se{and thus foster trust (which may lead to more widespread adoption of the metric)}. 
    Exemplary works in this direction are \citet{fomicheva-2021} and \citet{kequin-etal-2023-towards}. Accessible metrics might also help with use cases like translation error correction in the downstream tasks of language learning \revision{(non-experts)} or human translation services \revision{(translation experts)} \citep[e.g.,][]{zouhar-etal-2023-poor}. In these latter two use cases, \revision{a high average explanation \textit{plausibility} (to translation experts) and a strong metric performance are important. This is because, in cases where (1) the metric output is correct and (2) the explanation of the output is plausible to an expert, the explanation very likely correctly describes the shortcomings and strengths of the translation (on the level of an expert).\footnote{\revision{To foster trust, it might be sufficient, but unethical, if explanations of any metric (also weak ones) are plausible to its users (not necessarily to experts) \citep[e.g.,][]{jacovi-2020, Jin2023RethinkingAE}. The problem is that users could be tricked to believe into the performance of weak metrics or adapt the mindset of wrong explanations that have believable arguments. Hence, benchmarks of metrics and explanations in their correlation to human expert quality scores and explanations are important. Generally, users should be informed that hallucinations, where an explanation is untrustworthy, might occur.}} \textit{Faithfulness} is not necessary in this scenario, because the quality of the explanation in describing the translation already follows from (1) and (2). However, faithfulness is still desirable, as the internal decision process of a strong metric might support giving correct explanations (see \S\ref{sec:explainability}).} 
    \se{Goal 2} is important for the audience of \cliii{metric} users. 
    These could, for example, be non-experts that want to understand specific \cliii{metric} outputs or developers of MT systems that want to understand if a metric is a good fit for their tasks (see Figure \ref{fig:MTUseCases}).
    \item \textbf{\se{Goal 3}: Support Semi-Automatic labeling.} Fine-grained (human) annotations like word-level translation error labels \citep[e.g.,][]{fomicheva-2021} are \se{e}specially difficult to obtain. Obtaining automatic explanations to aid human annotators could boost their efficiency \se{tremendously} \cl{\cite[e.g.,][]{pmlr-v133-desmond21a}}. To our \se{best} knowledge, none of the current works has applied explainers in these settings for metrics. Semi-automatic labeling could also help with post-editing of translations \citep[e.g.,][]{specia-farzindar-2010-estimating} to guide how sentences should be adapted.
    \se{Goal 3}
    is important for metric developers that want to create \se{such word-level translation error data sets and metric users \revision{who} want to correct translations}. The requirements for \textit{faithfulness} and \textit{plausibility} are the same as for goal 2.
    \item \textbf{\se{Goal 4}: Checking for social biases.} Learned metrics might be biased towards certain types of texts used during training \citep[e.g.,][]{sun-etal-2022-bertscore}. Biases could be detected by observing explanations where sensitive attributes (for example, gender, political or racial aspects) are flagged to be important  \citep{lipton-2016}. For example, explanations could show that a metric considers male names more important for a translation than female names in the same scenario.  
    \se{Goal 4} \cl{can be important for all audiences.} 
    Affected users might be treated unfairly if biases occur in a metric. Metric users, metric developers, regulatory entities and managers might want to prevent this from happening to conform with given regulations. The requirements for \textit{faithfulness} and \textit{plausibility} are the same as for goal 1.
\end{itemize}

\section{\clii{Literature Review \& Selection}}
\label{selection}
To build the taxonomy in \S\ref{sec:taxonomy}, we survey literature which either explicitly state that they address the explainability of MT metrics or propose techniques that make MT metrics or their outputs clearer to understand by certain audiences (techniques that provide explanations). We restrict \cliii{our search and taxonomy} to works that consider neural network based MT metrics.
We conduct\se{ed} an initial literature search by selecting 11 ``seed papers''.
\cliii{Then, we leveraged the API of Semantic Scholar\footnote{\url{https://www.semanticscholar.org/}} to query new papers citing
the work of the seed papers and papers citing those papers (citation tree at depth 2). This amounted to 383 papers. Next, we filtered the papers by keywords, leaving 177 papers. Then, we manually annotated whether their contents fit our definitions. Thereby, we identified 12 papers that consider the explainability of MT metrics. In later stages, we progressively added further work released after the initial search and related works that we identified from other sources. In total, we describe 37 papers, some of which are not shown in Figure \ref{fig:taxonomy} due to space constraints, but discussed throughout the chapter.}

Following \S\ref{sec:metrics}, we consider higher granularity scores (e.g., word-level) as explanations for lower granularity metrics (e.g., sentence-level), i.e., \textit{feature importance} explanations. \revision{Due to the high number of existing word-level approaches,} for this explanation type we only consider works that explicitly employ word-level scores to explain sentence-level scores. 

\cliii{Likewise,} following \S\ref{sec:metrics}, metrics that report multiple \textit{quality aspects} such as adequacy or fluency provide more explanatory value \cliii{on why a translation is good or bad} than metrics that return a single score. For the \textit{quality aspect} explanation type, we make an exemplary selection of techniques derived from recent metrics that leverage generation probabilities \citep[e.g.,][]{thompson-2020,yuan-etal-2021-advances}. 

We also include a few non-MT works into our taxonomy, which show how some aspects that are currently not addressed by the MT related works might be tackled. 

\begin{figure}
    \centering
    \resizebox{\textwidth}{!}{\begin{tikzpicture}[
    every text node part/.style={align=center},
    category/.style={rectangle, fill=white, anchor=center, xshift=-0.5mm},
    title/.style={rectangle, fill=black!2, anchor=center},
    content/.style={rectangle, fill=black, opacity=1, fill opacity=0.06, text opacity=1, anchor=center},
    evalbox/.style={rectangle, fill=black, opacity=1, fill opacity=0, text opacity=1, anchor=center, draw=pink},
    >={Latex[width=1.5mm,length=1.5mm]},
]
\useasboundingbox (-3.2,0) rectangle (21.5,21.5);
\large
\pgfmathsetmacro{\tmargin}{0.1}
\pgfmathsetmacro{\tboxwidth}{5.1}
\pgfmathsetmacro{\tboxwidthb}{2.8}
\pgfmathsetmacro{\titleboxwidth}{4.3}
\pgfmathsetmacro{\tboxheight}{0.985}
\pgfmathsetmacro{\tcategoryheight}{0.3}
\pgfmathsetmacro{\tcategoryrelfontsize}{0}

\newcommand{\rAs}{\rBe + \tmargin} \newcommand{\rAe}{\rAs + \tboxheight}
\newcommand{\rBs}{\rCe + \tmargin} \newcommand{\rBe}{\rBs + \tboxheight}
\newcommand{\rCs}{\rDe + \tmargin} \newcommand{\rCe}{\rCs + \tboxheight}
\newcommand{\rDs}{\rEe + \tmargin} \newcommand{\rDe}{\rDs + \tboxheight}
\newcommand{\rEs}{\rFe + \tmargin} \newcommand{\rEe}{\rEs + \tboxheight}
\newcommand{\rFs}{\rGe + \tmargin+ \tcategoryheight} \newcommand{\rFe}{\rFs + \tboxheight}
\newcommand{\rGs}{\rHe + \tmargin} \newcommand{\rGe}{\rGs + \tboxheight}
\newcommand{\rHs}{\rIe + \tmargin} \newcommand{\rHe}{\rHs + \tboxheight}
\newcommand{\rIs}{\rJe + \tmargin} \newcommand{\rIe}{\rIs + \tboxheight}
\newcommand{\rJs}{\rKe + \tmargin} \newcommand{\rJe}{\rJs + \tboxheight}
\newcommand{\rKs}{\tmargin} \newcommand{\rKe}{\rKs + \tboxheight}

\newcommand{\cAs}{\tmargin} \newcommand{\cAe}{\cAs + \tboxwidth}
\newcommand{\cBs}{\cAe + \tmargin} \newcommand{\cBe}{\cBs + \tboxwidthb}
\newcommand{\cCs}{\cBe + \tmargin} \newcommand{\cCe}{\cCs + \tboxwidthb}
\newcommand{\cDs}{\cCe + \tmargin} \newcommand{\cDe}{\cDs + \tboxwidth}
\newcommand{\cEs}{\cDe + \tmargin + \tmargin} \newcommand{\cEe}{\cEs + \tboxwidth}

\newcommand{\rtitle}[2]{
    \fill[title] (-\tmargin-\titleboxwidth/1.75, \csname r#1s\endcsname) rectangle node{\vphantom{$^i_ a$}#2\vphantom{$^i_a$}} (-\tmargin, \csname r#1e\endcsname);
}

\newcommand{\rtitleb}[3]{
    \fill[title] (-\tmargin-\titleboxwidth/1.75, \csname r#1s\endcsname + \tboxheight) rectangle node{\vphantom{$^i_ a$}#3\vphantom{$^i_a$}} (-\tmargin, \csname r#2e\endcsname - \tboxheight);
}

\newcommand{\rcategory}[2]{
    \fill[category] (-\tmargin-\tboxwidth, \csname r#1e\endcsname + \tmargin + \tcategoryheight) rectangle node{\relsize{\tcategoryrelfontsize} #2} (-\tmargin, \csname r#1e\endcsname + \tmargin);
    \draw (\cAs, \csname r#1e\endcsname + \tmargin) -- (\cDe, \csname r#1e\endcsname + \tmargin);
    \draw (\cEs, \csname r#1e\endcsname + \tmargin) -- (\cEe, \csname r#1e\endcsname + \tmargin);
}

\newcommand{\ctitle}[2]{
    \fill[title] (\csname c#1s\endcsname, \rAe + \tmargin + \tmargin) rectangle node{\vphantom{$^i_ a$}#2\vphantom{$^i_a$}} (\csname c#1e\endcsname, \rAe + \tmargin + \tmargin + \tboxheight);
}

\newcommand{\ccategory}[2]{
    \fill[category] (\csname c#1s\endcsname - \tboxwidth/2, \rAe + \tmargin + \tmargin + \tboxheight + \tmargin + \tcategoryheight) rectangle node{\relsize{\tcategoryrelfontsize} #2} (\csname c#1e\endcsname, \rAe + \tmargin + \tmargin + \tboxheight + \tmargin);
    \draw (\csname c#1s\endcsname - \tmargin, \rJs) -- (\csname c#1s\endcsname - \tmargin, \rAe);
}

\newcommand{\tcontent}[4]{
    \fill[content] (\csname c#2s\endcsname, \csname r#1s\endcsname) rectangle node{\vphantom{$^i_ a$}#4\vphantom{$^i_a$}} (\csname c#3e\endcsname, \csname r#1e\endcsname);
}

\newcommand{\tcontentb}[5]{
    \fill[content] (\csname c#3s\endcsname, \csname r#1s\endcsname +\tboxheight) rectangle node{\vphantom{$^i_ a$}#5\vphantom{$^i_a$}} (\csname c#4e\endcsname, \csname r#2e\endcsname - \tboxheight);
}

\newcommand{\tcontentc}[5]{
    \fill[evalbox] (\csname c#3s\endcsname, \csname r#1s\endcsname +\tboxheight) rectangle node{\vphantom{$^i_ a$}#5\vphantom{$^i_a$}} (\csname c#4e\endcsname, \csname r#2e\endcsname - \tboxheight);
}

\draw[-] (\cAs, \rAe + \tmargin + \tmargin + \tboxheight + \tcategoryheight + \tmargin + \tmargin) -- (\cEe, \rAe + \tmargin + \tmargin + \tboxheight + \tcategoryheight + \tmargin + \tmargin ) node[above left, xshift=-0.2cm]{less information \hspace{17.9cm} more information};

\rcategory{A}{}
\rtitleb{A}{B}{quality\\aspects}
\rtitleb{C}{E}{feature\\importance}
\rtitle{F}{fine-grained\\errors}
\rcategory{G}{}
\rtitleb{G}{I}{\cliii{perturbation}\\robustness}
\rtitle{J}{linguistic\\properties}

\ccategory{A}{post-hoc \hspace{0.9cm}}
\ctitle{A}{black-box}
\ctitle{B}{gradient}
\ctitle{C}{attention\\weights}
\ctitle{D}{embeddings}
\ccategory{E}{intrinsic \hspace{1.1cm}}
\ctitle{E}{model specific}

\tcontentb{A}{B}{E}{E}{\citep{thompson-2020}\\\citep{yuan-etal-2021-advances}\\\citep{eddine-etal-2022-datscore}\\\citep{jinlan-etal-2023-gptscore}}
\tcontent{A}{A}{A}{\citep{kaster-etal-2021-global}$^E$}
\tcontent{A}{D}{D}{\citep{golovneva2023roscoe}$^N$\\\citep{opitz-frank-2022-sbert}$^N$}
\tcontent{D}{B}{C}{\citep{rei-etal-2022-cometkiwi}$^E$\\\citep{Rei2023TheIS}$^E$}
\tcontent{E}{A}{C}{\citep{treviso-etal-2021-ist}$^E$}
\tcontentb{C}{E}{E}{E}{\citep{fernandes-etal-2022-learning}$^E$\\\citep{kabir-carpuat-2021-umd}$^E$\\\citep{rubino-2021}$^E$\\\citep{wang-etal-2022-fine}$^E$}
\tcontentb{D}{E}{D}{D}{\citep{leiter-2021-reference}$^E$\\\citep{tao-etal-2022-crossqe}$^E$\\\citep{fatemeh-etal-2022-mismatching}$^E$}
\tcontent{D}{A}{B}{\citep{eksi-etal-2021-explaining}$^E$}
\tcontent{C}{A}{D}{\citep{fomicheva-etal-2022-translation}$^E$}
\tcontent{C}{A}{A}{\citep{BMX}}
\tcontent{F}{E}{E}{\citep{lu2022humanlike,lu2023error}\\\citep{xu2023instructscore}$^E$, ...}
\tcontentb{G}{I}{A}{A}{\citep{sai-2020}$^E$\\\citep{karpinska-etal-2022-demetr}$^E$\\\citep{he-etal-2022-on}$^R$\\\citep{sun-etal-2022-bertscore}$^F$\\\citep{chen2022menli}$^R$\\...}
\tcontent{J}{A}{A}{\citep{kaster-etal-2021-global}$^E$}
\tcontent{J}{D}{D}{\citep{opitz-frank-2022-sbert}$^N$}

\draw[-] (\cAs - \tmargin  - \tboxwidth/1.75 + 0.3, \rAe) -- (\cAs - \tmargin - \tboxwidth/1.75 - \tmargin +0.3, \rJs) node[anchor=mid, below, rotate=-90, xshift=-6cm]{\;\;\;\;\;\;\;\cl{decision-understanding} \hspace{2.7cm} \cl{model-understanding}};

\end{tikzpicture}
}
    \caption{Overview of papers addressing the explainability of machine translation evaluation. The graphic structure is adapted from a survey on post-hoc methods for explainable NLP by \citet{madsen-etal-2022-post}. The rows show the types of explanations returned by the respective methods. They are ordered from \cl{decision-understanding} to \cl{model-understanding}. \cl{Decision-understanding} techniques explain specific outputs of a \cliii{metric}, while \cl{model-understanding} techniques describe general properties. The columns show the required level of model access that each explainability technique needs. Fields, where boxes overlap are colored in a darker \clii{grey}. \cliii{Note that most work listed under feature importance was developed as part of the Eval4NLP21} \citep{fomicheva-2021} and the WMT22 QE \citep{zerva-EtAl-2022-WMT} shared tasks. $^N$ indicates that a technique is not directly from the MT domain. $^R$ indicates that a technique was proposed to tackle robustness. $^F$ indicates that the technique was proposed to explore fairness aspects. $^E$ is set when the papers' authors themselves perceived that their method tackles explainability.}
    \label{fig:taxonomy}
\end{figure}

\begin{figure}[htb!]
    \centering
    \resizebox{\textwidth}{!}{\begin{tikzpicture} [set/.style = {draw}]
\node (A)[set, align=left, rectangle, line width = 0.7mm] at (0, 5.2) {\textbf{Quality Aspects}\\Adequacy: 0.7\\Fluency: 1};
\node (B)[set, align=left, rectangle, line width = 0.7mm] at (-6, 3) {\textbf{Feature Importance}\\ $0.5$ $0.1$ $0.5$ $\shortrightarrow$ ``\green{Things}  \red{became} \green{complicated}''\\ $0.5$ $0.4$ $0.1$ $0.5$ $\shortrightarrow$ ``\green{Die Dinge} \red{bekamen} \green{kompliziert}''};
\node (C)[set, align=left, rectangle, line width = 0.7mm] at (6, 3) {\textbf{Fine Grained Errors}\\``Things $<$sense$>$\red{became}$<$/sense$>$ complicated''\\``Die Dinge $<$sense$>$\red{bekamen}$<$/sense$>$ kompliziert''};
\draw [dotted] (-10,0) -- (10,0);
\node (D)[set, align=left, rectangle, line width = 0.7mm] at (-6,-3) {\textbf{\cliii{Perturbation} Robustness}\\
Numeric: ``3 Dinge bekamen kompliziert''$\shortrightarrow$lower score\\Order: ``kompliziert die Dinge bekamen''$\shortrightarrow$lower score\\...};
\node (E)[set, align=left, rectangle, line width = 0.7mm] at (6,-3) {\textbf{Linguistic Properties}\\\includegraphics[width=0.36\textwidth]{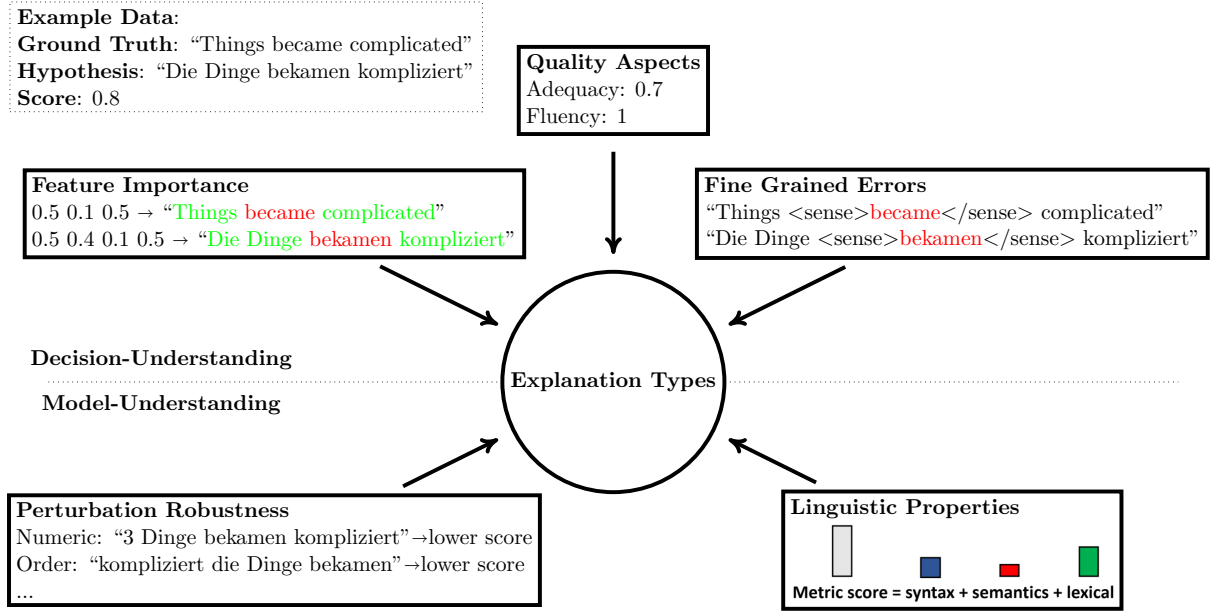}};

\node (Example)[set, align=left, rectangle, dotted] at (-6.5, 5.8) {\textbf{Example Data}:\\
\textbf{Ground Truth}: ``Things became complicated''\\
\textbf{Hypothesis}: ``Die Dinge bekamen kompliziert''\\
\textbf{Score}: $0.8$};

\node (G)[align=left] at (-8, 0.4) {\textbf{Decision-Understanding}};
\node (H)[align=left] at (-8, -0.4) {\textbf{Model-Understanding}};
\node (F)[set, align=left, circle, line width = 0.7mm, fill=white] at (0,0) {\textbf{Explanation Types}};

\draw[{[sep=6.2pt]}-{Computer Modern Rightarrow[length=3mm, width=3mm]}, line width=0.7mm, shorten <=8pt, shorten >=8pt]
(A) edge (F) (B) edge (F) (C) edge (F) (D) edge (F) (E) edge (F) ;
\end{tikzpicture}}
    \caption{Explanation types shown in \S\ref{fig:taxonomy}. \cl{For each type we provide a hypothetical example. For this we assume the exemplary translation that is shown in the top-left box. Here, the polysemic word "became" is translated with the wrong word sense "to receive" instead of "to get/to develop into". The correct translation is "wurden". The example for \textit{perturbation robustness}, perturbs the hypothesis sentence. 
    }}
\label{fig:exTypes}
\end{figure}

\section{Taxonomy}
\label{sec:taxonomy}
Based on the background described in the previous \se{sections} and the literature search we conducted, we create a taxonomy of prior works that consider the explainability of MT metrics. This taxonomy is shown in Figure \ref{fig:taxonomy}. The taxonomy contains \se{five} explanation types based on the selected papers\cl{'} contents: \textit{quality aspects}, \textit{feature importance}, \textit{fine-grained errors}, \textit{\cliii{perturbation} robustness} and \textit{linguistic properties}.  We structure this section based on these explanation types. Figure \ref{fig:exTypes} shows an example for each explanation type. For each type, we describe the techniques applied by the previous works and how they relate to each other. We begin with explanation types that explain specific samples and move to techniques that explain metric behavior in general. 

\noindent\textit{\\Quality aspects \cliii{(Row 1 in Figure \ref{fig:taxonomy})}} \;\;\;\; This category comprises works that present explanations of the metric output by producing separate scores for different aspects of the translation. \cl{Most} methods that we list develop \cl{generation-based} metrics that are explainable by design \cl{(``model specific'' in Figure \ref{fig:taxonomy})}. \cliii{This means},\ \cl{they do not use} an external explainer that could be applied to other metrics. BARTScore \citep{yuan-etal-2021-advances} uses the autoregressive generation model BART \citep{lewis-2020} to predict the average word generation probability of a sentence \cliii{$B=(b_1,...,b_m)$}, given a sentence \cliii{$A=(a_1,...,a_n)$}:
$$BARTScore(A\longrightarrow B)=\sum_{t=1}^m \log p(b_t|b_{<t},A,\theta)$$
\cliii{This formula sums over the log-probability of token $b_t$ being generated by BART with parameters $\theta$ given all previous tokens of $B$ and all tokens of $A$.} Depending on whether A is the ground truth sentence and B the hypothesis (or vice-versa), \citet{yuan-etal-2021-advances} state that different aspects of the translation are evaluated. \cliii{For example}, they associate the generation direction of A:source $\longrightarrow$ B:hypothesis with the properties coherence and fluency. This idea of using the generation probabilities in multiple directions was first introduced in PRISM \citep{thompson-2020}, but \citet{yuan-etal-2021-advances} 
formalized the connection of the directions to known quality \cl{aspects}. DATScore \citep{eddine-etal-2022-datscore} extends BARTScore by using a multilingual model and further generation directions (\cl{additionally} using the source sentence). 
The metric GPTScore \citep{jinlan-etal-2023-gptscore} provides a suite of prompts for language models that emphasize specific aspects when requesting a translation \cl{or paraphrase}. Then, similar to BARTScore, the generation probabilities are averaged over all tokens. \cliii{For example}, they use the prompt ``Rewrite the following text to make it more grammatical and well-written: {\cl{[reference or hypothesis]}} In other words,''  
to \cl{generate a paraphrase} whose probabilities they use for a fluency score. \cliii{The example specifies \textit{reference/hypothesis} to describe that the metric checks either the generation probability of the hypothesis given the reference, or vice-versa.} \\
Another way to gather fine-grained scores is to apply suites of metrics (each of which considers \cliii{different} aspects), \cliii{like} ROSCOE \cite{golovneva2023roscoe}, \cl{an embedding based metric suite}, which has been proposed to grade step-by-step reasoning processes \cl{(column ``embeddings'' in Figure \ref{fig:taxonomy})}. \\
\cliii{Besides adding explanatory value per se, quality aspect explanations have two more potential use cases for explainable metrics: (1) separate quality scores could be aggregated, to compute single, decomposable sentence-level scores and (2) scores could be aggregated over data sets to check which aspects a metric favors the most. For example, \citet{kaster-etal-2021-global} and \citet{opitz-frank-2022-sbert} also assign quality aspect scores, but mostly use them for model understanding}; \se{see} \textit{linguistic properties} explanations for more details. \\
In the example presented in Figure \ref{fig:exTypes}, we report fluency and adequacy as quality aspects. We assign a fluency score of 1 \se{on a scale from 0 to 1}, as the translation is fluent and an adequacy of 0.7 as the translation of ``became'' is incorrect. 

\limit{\revision{\textbf{Explanation Type:} Quality Aspects}\\\textbf{Goals:} \textit{Quality aspect} approaches could be used for metric accessibility and semi-automatic labeling. The usage for metric diagnosis is limited, as the approaches are \clii{often} based on models that are explainable by design and might not be faithful (the models generate their own explanations). The current approaches cannot be used to detect social biases.\\\textbf{Audiences:} Quality aspects like \textit{fluency} and \textit{adequacy} follow simple definitions, such that these metrics could be used by non-expert and expert users.\\\textbf{Limitations:} The generation directions might not fully match the quality aspects. Also, the usage of prompting leaves it to another black box setting whether the model picks up the desired properties or not. Lastly, the explanatory value of quality aspect approaches \cliii{per se} is rather small, as they \cliii{only} allow to reason, ``the rated translation is good because of aspect x'' and to guess about the model's reasoning. \cliii{As none of the current works\footnote{\citet{kaster-etal-2021-global} and \citet{opitz-frank-2022-sbert} are considered at a later point.} specifically targets explainability, they do not consider faithfulness. Plausibility is automatically considered in standard evaluation schemes with human judgments.}} 

\noindent\textit{\\Feature importance \cliii{(Row 2 in Figure \ref{fig:taxonomy})}} \;\;\;\; \label{sec:featureAttribution}
Feature importance techniques assign scores to all features in the input of a \cliii{metric} (here the input tokens) that express their importance to the output of the \cliii{metric}.\footnote{Other names include \textit{relevance scores} or \textit{attribution scores}.} 
\citet{fomicheva-etal-2022-translation} note that \cliii{high} feature importance scores \cliii{should} correspond to translation errors, \cliii{as humans pay most attention to errors when rating a translation}. They build on the idea that when humans evaluate translations, they often focus on the errors that they can identify on a word- or phrase-level \citep{freitag-etal-2021-experts}. Then, \citet{fomicheva-etal-2022-translation} evaluate how well feature importance is correlated with human word-level error annotations, i.e., the plausibility of the explanations.
 
\cliii{We note that, depending on the explainability method, these scores might be inverted, following the interpretation ``errors are not important for a metric to achieve a high score''.}\footnote{\cliii{Most gradient and attention based approaches return high scores for errors \citep[e.g.,][]{treviso-etal-2021-ist, rubino-2021}, for embedding-based approaches it depends on whether cosine-similarity or -distance is used \citep[e.g.,][]{leiter-2021-reference, tao-etal-2022-crossqe}, for attribution based techniques using LIME \citep{ribeiro-2016} or SHAP \citep{lundberg-2017}, low scores are returned for errors (e.g., as part of the baseline implementations in Eval4NLP21: \url{https://github.com/eval4nlp/SharedTask2021/tree/main/baselines)}}.}
Two shared tasks for the explainability of reference-free metrics have originated from the setting \cliii{in \citet{fomicheva-etal-2022-translation}} so far \citep{fomicheva-2021, zerva-EtAl-2022-WMT}. As the respective papers already give an overview of the participating methods, we will only briefly describe the methods and set our focus to describe the model access they require. \cl{We describe the model access levels of Figure \ref{fig:taxonomy} from left to right.} Approaches that work directly on \cl{black box models} are explored in \citet{fomicheva-etal-2022-translation}, \citet{eksi-etal-2021-explaining} and \citet{treviso-etal-2021-ist}. They leverage model agnostic explainers such as LIME \citep{ribeiro-2016} and SHAP \citep{lundberg-2017}. These methods check how a metric's sentence-level score changes when the input sentence is \cliii{perturbed} at selected positions. While the results correlate with the human annotations to some degree, techniques that use model access performed better in the shared tasks. Gradient based methods are explored by \citet{fomicheva-etal-2022-translation}, \citet{eksi-etal-2021-explaining}, \citet{treviso-etal-2021-ist} and \citet{rei-etal-2022-comet}. These papers use methods like Integrated Gradients \citep{sundararajan-2017} to determine the effect each input token has on the output. Attention values perform especially good for \citet{treviso-etal-2021-ist} and \citet{rei-etal-2022-cometkiwi}, leading to winning techniques for most language pairs in both shared tasks, \cliii{i.e., producing the most plausible explanations. In specific, \citet{treviso-etal-2021-ist} succeed by scaling attention weights by the norm of value vectors and filtering for attention heads that have the highest scores in detecting errors. \citet{rei-etal-2022-cometkiwi} build on their approach and (1) instead scale attention weights with the l2-norm of the gradient of value vectors and (2) designing an ensemble method for attention heads. In their final submission, they ensemble the best performing attention heads over all layers.}
Word-level contextualized embeddings are used by \citet{leiter-2021-reference}, \citet{tao-etal-2022-crossqe} and \citet{fatemeh-etal-2022-mismatching}. Specific\se{ally}, they leverage the word-level embedding distances that are inherently computed in token-matching metrics (see \S\ref{sec:metrics}) and use them as importance scores. For model-specific works,  \citet{rubino-2021} jointly fine-tune their metrics on the word-level and sentence-level using synthetic data, to directly output the computed word-level scores.\footnote{\citet{rubino-2021} won the constrained track of Eval4NLP21, which did not allow for word-level finetuning.} \citet{kabir-carpuat-2021-umd} use LIME with Transquest \citep{ranasinghe-etal-2020-transquest-wmt2020} and merge the word-level importance scores with additional word-level scores obtained from a pure word-level model. 
After the shared tasks, \citet{fernandes-etal-2022-learning} enhance explainability techniques (in case of MT they use a gradient technique) by adding learnable parameters and employing a student teacher mechanism to improve the correlation to human word-level scores.
\cliii{\citet{Rei2023TheIS} extend the work of \citet{rei-etal-2022-cometkiwi} by using reference-based metrics and importance scores of the references. They construct a data set from the WMT21 MQM annotations \cite{freitag-etal-2021-results} by assuming that feature attribution scores link to annotated error spans. Evaluations on this data set show that the winning explainability techniques of the shared task perform best and that references can improve the plausibility. Notably, they also show that explanations can help to detect critical errors in translations. Based on this, they employ the explanations for metric diagnosis, to test whether their metrics are more prone to ignoring certain types of critical errors. In their limitations, they note that the explainability techniques they use have shown to be faithful in some scenarios, but also note that they did not specifically test whether they are faithful for their architecture.}
\cliii{Finally, we note that many more metrics can provide feature-importance like explanations by design. Similar to the base metrics used by \citet{leiter-2021-reference}, \citet{tao-etal-2022-crossqe} and \citet{fatemeh-etal-2022-mismatching}, most unsupervised matching metrics could be used to determine importance scores based on word- or embedding-similarities. Further, the per-word generation probabilities in text generation metrics like PRISM or BARTScore can also be interpreted as feature importance (they can also be interpreted as the confidence of the generation models; \citealt{fomicheva-2020b}).}\\
\cliii{In the example presented in Figure \ref{fig:exTypes}, high feature importance scores are assigned to words that were translated correctly, i.e., ``Things/complicated'' and lower scores indicate words that are wrong.} \\
\cl{\citet{BMX} take a different approach on using feature importance techniques for the goal of metric improvement. Instead of using the \cliii{feature importance} scores in a task that might involve human explainees, the authors aggregate and directly integrate \cliii{the feature importance scores} into metric scores. This follows the intuition that the explanations carry additional information that can be beneficial to the sentence-level correlation of a metric with human judgments.}

\limit{\revision{\textbf{Explanation Type:} Feature Importance}\\
\textbf{Goals:} \cliii{The discussed \textit{feature importance} approaches are tested for plausibility}, so strong approaches could be used for metric accessibility/improvement and semi-automatic labeling. The usage for metric diagnosis is currently limited, as only samples are explained, and \cliii{approaches have not been tested for faithfulness}. The current approaches have not been used to detect social biases.\\\textbf{Audiences:} The relation of low/\cliii{high} scores and translation errors is intuitive, such that these techniques could be used by non-expert and expert users. Metric developers could use the methods to create new data sets \revision{(by supporting the fine-grained human annotation of errors)} or to directly incorporate them into metrics \cite{BMX}.\\\textbf{Limitations:} For all approaches, \cliii{it is not specifically tested whether the explanations are faithful.}\footnote{In different possible aspects \citep{jacovi-2020}} Notably, \citet{Rei2023TheIS} consider that faithfulness is probably given in their case but note the missing tests as limitation.
Second, certain translation errors cannot be easily captured by highlighting specific words. For example, the word-level scores cannot represent cases where the MT fails to explicitly express a grammatical feature that is omitted in the source language but is required to be explicit in the target language (for example,\ the use of articles when translating from Russian into German).
Third, the explanation type does not provide correspondence between highlighted error words in the source and target language.}

\noindent\textit{\\Fine-grained errors \cliii{(Row 3 in Figure \ref{fig:taxonomy})}} \;\;\;\; Some recent works detect errors on a word-level and leverage their severity, number and type to compute a sentence-level score \citep{lu2022humanlike, lu2023error, kequin-etal-2023-towards,xu2023instructscore, fernandes-etal-2023-devil, kocmi-federmann-2023-gemba}. This follows the ideas of the MQM annotation scheme \citep{lommel-2014, freitag-etal-2021-experts} in which human \revision{translation} experts evaluate translations in a similar manner. \citet{freitag-etal-2021-experts} have shown that this annotation scheme improves over previous annotation methods such as direct assessments \citep{graham-2015}. While the main goal of these works is to provide sentence-level scores, \cl{they are also explainable-by-design and the generated error-labels carry explanatory value (model-specific in Figure \ref{fig:taxonomy})}. In specific,  BARTScore++ \citep{lu2022humanlike} distinguishes explicit errors that are easy to detect and implicit errors. They detect the explicit errors based on BART generation probabilities. Then they generate a sentence-level score based on an additional improved hypothesis sentence in multiple BARTScore calculations. 
\citet{lu2023error} also combine fine-grained errors into a score. They first request ChatGPT to generate explanations based on MQM annotations (erroneous words, error type and error majority). In a second step, they ask a model to calculate a score from these. \revision{This is a form of chain-of-thought prompting \citep{wei2022chain}, a prompting technique where a language model is instructed to first generate an explanation/reasoning before returning a result}. \revision{\citet{lu2023error} find that their approach is competitive to other state-of-the-art metrics, especially with newer versions of ChatGPT}. 
\citet{xu2023instructscore} fine-tune InstructScore, a LLaMA  model \citep{llama} that generates a text containing error majority, error spans, error types (MQM) and an explanation for every error in an input sentence. The generated texts are similar to \citet{lu2023error} but add more details on each error. Similarly,  \citet{xu2023instructscore} construct a score from the count and severity of errors. They obtain synthetic training data from GPT4 \citep{gpt4}, analyze failure modes of InstructScore during training and later optimize the metric on data sets with real machine translation outputs and annotated ground truth scores. Their model shows state-of-the-art results for reference-free sentence-level evaluations. Further, the detailed error annotations can be used as fine-grained explanations. \revision{Contrary to Instructscore, instead of fine-tuning, \citet{fernandes-etal-2023-devil} propose AutoMQM a prompting technique that leverages few-shot learning \citep{NEURIPS2020_1457c0d6} and chain-of-thought prompting to generate MQM error annotations. They use it to prompt PALM 2 \citep{anil2023palm} and achieve competitive results for system- and sentence-level evaluation. In a similar way, \citet{kocmi-federmann-2023-gemba} design the \textit{GEMBA-MQM} prompting technique and prompt GPT4 in a few-shot and chain-of-thought setting to produce MQM annotations. A main distinction is that AutoMQM dynamically selects few-shot examples from pools of example translations and ratings based on the input. In contrast, GEMBA-MQM uses only a few, pre-selected examples and is therefore easier to apply to further language pairs. Both, AutoMQM and GEMBA-MQM calculate their final scores based on the detected errors with MQM heuristics \citep{freitag-etal-2021-experts}.} \\
The WMT 2023 shared task on quality estimation \citep{blain-etal-2023-findings} also includes a new sub-track where the submitted metrics should detect fine-grained errors and their severity on MQM data, to facilitate the development of more explainable metrics. The participants explore two main directions: (1) detection with GPT4 \citep{rei-etal-2023-scaling}\footnote{Another participating team \textit{KUNMT} did not submit a paper.} and (2) fine-tuning models for error span detection, either based on word-level supervision \citep{geng-etal-2023-unify,li-etal-2023-hw-tsc} or with multi-task objectives \citep{rei-etal-2023-scaling}. Notably the two winning submissions from \citet{geng-etal-2023-unify} and \citet{rei-etal-2023-scaling} are ensemble approaches, where \citet{rei-etal-2023-scaling} also use pseudo-references from Deepl and Google Translate (if their quality is better than a threshold, as measured by another metric) \citep{blain-etal-2023-findings}. \\
Similarly, but not related to the shared task, \citet{kequin-etal-2023-towards} fine-tune a seq2seq model to predict addition and omission errors on a word-level. \\
\cl{In the example presented in Figure \ref{fig:exTypes}, a tag ``sense'' flags the wrong word sense of ``became''}. Notably, all \textit{fine-grained error} approaches we describe are very recent, which is potentially caused by the adaptation of MQM in the WMT metrics shared tasks and the recent successes of LLMs.

\limit{\revision{\textbf{Explanation Type:} Fine-grained errors}\\\textbf{Goals:} \cliii{These \textit{fine-grained error} approaches can be used for increased accessibility and semi-supervised labeling, as the methods model plausible explanations (some of them by testing the alignment of scores aggregated from explanations with human MQM scores)}. The usage for metric diagnosis is limited, as only samples are explained, and explanations have not been tested for faithfulness. The current approaches cannot be used to detect social biases.\\\textbf{Audiences:} Flagged translation errors are simple to understand, such that these techniques could be used by non-expert and expert users. Metric developers might use them to create new data sets to train new metrics.\\\textbf{Limitations:} The current approaches (besides \citealt{xu2023instructscore}) do not have the direct goal to provide explanations that make the sentence-level scores more descriptive. Instead, they aim to build sentence-level metrics from LLM outputs. As such, there might be open opportunities in leveraging the explanations provided by these models to explain to a human explainee. Another limitation are missing evaluations of faithfulness.}

\noindent\textit{\\\cliii{Perturbation} robustness \cliii{(Row 4 in Figure \ref{fig:taxonomy})}} \;\;\;\;
The methods described up to now addressed \cl{decision-understanding} techniques \clii{\cite{Gehrmann2020}}. This paragraph and the next describe \cl{model-understanding} techniques (we \clii{describe} these terms in \S\ref{sec:explainability}). 
Many recent works evaluate the robustness of MT metrics \citep{sai-2021, karpinska-etal-2022-demetr, he-etal-2022-on, sun-etal-2022-bertscore, chen2022menli, vu-etal-2022-layer}. \clii{Robustness evaluation qualifies as \cl{model-understanding}, as it usually infers properties that describe a model} (see \S\ref{sec:explainability}). All works in this category are structured into ``black box'' in Figure \ref{fig:taxonomy}.

\citet{sai-2021} propose perturbation checklists, \se{following \citet{ribeiro-etal-2020-beyond}},  that evaluate how susceptible metrics are towards predefined types of \cliii{perturbations} as evaluation criterion. This allows developers to check whether all invariances that are required for a specific task are fulfilled. 
In particular, they compare the change in the metric's score with the change in score a human would assign after the perturbation.  
These templates encompass dropping or adding of context as well as negations. They show that perturbation checklists allow to pick metrics that are strong with respect to specific properties. For example, \se{their experiments indicate} MoverScore \citep{zhao-2019} would capture fluency better than BERTScore \citep{zhang-2020} due to its ability to cope with the jumbling of words. The checklists allow for more fine-grained assessment of automatic evaluation metrics, which exposes their limitations and supports the design of better metrics. 
\citet{chen2022menli} create a similar, but fully automatic variant of this evaluation. They design several \cliii{perturbation} types (framed as adversarial attacks) such as addition, omission and negation. Then they measure whether a tested metric prefers \clii{a meaning preserving, but lexically dissimilar paraphrase of a source/reference sentence over a minimally attacked but lexically similar version of the source/reference sentence}. \clii{This means they check} \cl{whether the metric score of the attacked example is successfully lower than the metric score of a paraphrase.} One of their findings is that standard metrics often have problems in detecting wrong names or numbers in translations. \citet{karpinska-etal-2022-demetr} create DEMETR, a data set of machine translations that were perturbed using several \cliii{perturbation} types. They propose to test a metric's preference between these \cliii{perturbations} and the original translation. \clii{Their task is easier for a metric to compute than \citet{chen2022menli}, as the preference is checked directly between attacked and non-attacked version, while \citet{chen2022menli} use a dissimilar paraphrase instead of the non-attacked version.} 
Also, they define more \cliii{perturbation} types than \citet{chen2022menli}, but require manual annotation for some of them. \citet{he-etal-2022-on} also compare the original translation with certain \cliii{perturbation} types. They differ from the other two papers in terms of domains they are applying their method to and in the \cliii{perturbation} types that they propose. \clii{\citet{vu-etal-2022-layer}} use character-level adversarial attacks to evaluate the robustness of several BERTScore variants. \\
The WMT21 \citep{freitag-etal-2021-results}, WMT22 \citep{freitag-etal-2022-results} and WMT23 \citep{freitag-etal-2023-results} metrics shared tasks introduce a similar setting for robustness tests. In WMT21, the organizers provided the participants with a challenge set composed of translations that were \cliii{perturbed} in a specific way, to test whether metrics reduce their score as expected when presented with respective phenomena. In WMT22 and WMT23, participants of a challenge set subtask \citep{alves-etal-2022-robust, pub12765, chen-etal-2022-exploring, amrhein-etal-2022-aces, lo-etal-2023-metric, amrhein-etal-2023-aces, avramidis-etal-2023-challenging} designed data sets posing challenges in a similar manner. \revision{In the example presented in Figure \ref{fig:exTypes}, we show tests like those performed by techniques of this category. Here, a \textit{model-understanding} explanation could be that the metric generally assigns lower scores when input sentences are subjected to numeric or word order perturbations.} \\
\cliii{The work by \citet{guerreiro-etal-2023-looking} is also related to these approaches. One of their contributions is an expert annotated data set of machine translation hallucinations. They use this data set to test how well selected metrics can identify each hallucination type. Further, they give the recommendation to simply use the sequence log-probability for hallucination detection.}
While the previous techniques describe the susceptibility of metrics towards various \cliii{perturbation} types, \citet{sun-etal-2022-bertscore} describe methods to analyze the metrics' social biases. They also base their evaluation on \cliii{perturbed} input sentences; however, they evaluate whether a metric grades two versions of a sentence that contain opposite social stereotypes the same with respect to a neutral reference sentence (that does not contain stereotypes). They find that generation-based metrics show the least biases.

\limit{\revision{\textbf{Explanation Type:} Perturbation Robustness}\\\textbf{Goals:} The presented \textit{perturbation robustness} approaches give detailed insights into metric weaknesses. Therefore, they are most suited to be used for metric diagnosis and improvement. \\\textbf{Audiences:} Metric developers can use these methods for diagnosis. They can also be used by metric users to check if a metric fits their use case. Regulatory entities might test if metrics fulfill their guidelines, \cl{e.g., exhibit no social biases}. Lastly, affected users might gather evidence when treated unfairly.\\\textbf{Limitations:} A drawback of the perturbation-based approaches is that they require predefined error types. Therefore, errors lacking a definition in the evaluation scheme that would fool the machine translation metrics could be missed. \cliii{These approaches are faithful to some degree, namely in the domains they have been evaluated in. For small perturbation sets it could be questionable if the model generally behaves as predicted. Therefore we would suggest evaluating each property on a large, general data sample. Still, it could be advisable to employ faithfulness tests, if possible, to achieve guarantees.}}  

\noindent\textit{\\Linguistic properties \cliii{(Row 5 in Figure \ref{fig:taxonomy})}} \;\;\;\;
Here, we summarize \cl{model-understanding} techniques that directly summarize various linguistic properties a metric fulfills. So far, the only method that falls into this category for MT is \citet{kaster-etal-2021-global} \cl{(see linguistic properties/black box in Figure \ref{fig:taxonomy})}. They propose a \cl{model-understanding} technique that decomposes the score of sentence-level BERT-based metrics into linguistic factors. In particular, they explore the degree to which metrics consider the properties \textit{syntax}, \textit{semantics}, \textit{morphology} and \textit{lexical overlap}. To do so, they measure these properties \se{in} a given data set and learn regressors that explain how much each distinct \se{property} contributes to the original metric's score. In their experiments, they show that each metric captures semantic similarity and lexical overlap to some degree. Syntactic and morphological similarity are either captured to a smaller extent or are even negatively correlated with the metric score. Especially, MoverScore \citep{zhao-2019} and BERTScore \citep{zhang-2020} have a comparatively high coefficient for the lexical score. \\
\revision{The example in} Figure \ref{fig:exTypes} \revision{shows that} these techniques can give explanations \revision{that highlight which properties a metric generally focuses on. In the specific example, the metric puts a high focus on lexical properties (green bar) and a low focus on semantics (red bar).} \\
Note that the explanation of linguistic properties is close to the \cl{decision-understanding} of quality aspects. If these are collected over many samples and a regression is performed, similar explanations might be achieved.
A related approach used for another NLP task is \se{by} \citet{opitz-frank-2022-sbert}, who retrain sentence embeddings to decompose into sub-components that match the aspects of abstract meaning representation, a format to represent the meaning of sentences. This allows them to attribute parts of the score to different aspects of a sentence. They evaluate their approach on the \cl{model-understanding} level. When this would be applied to MT, it would also count towards this explanation type.

We also list \citet{kaster-etal-2021-global} and \citet{opitz-frank-2022-sbert} with quality aspect explanations, as the metrics that measure the linguistic factors can also be computed on a sentence-level.

\limit{\revision{\textbf{Explanation Type:} Linguistic Properties}\\\textbf{Goals:} The discussed \textit{linguistic properties} approaches evaluate general \cliii{metric} properties and can be used for metric debugging. As they generate sentence/metric-level property scores, \clii{the techniques} \se{are} not usable for word-level semi-automatic labeling. When used on a sentence-level, \clii{the techniques} can aid the goal of making metrics more accessible. While not employed here, the technique\se{s} might be usable to detect social biases by introducing a set of bias detection scores from which to learn the regressors. 
\\\textbf{Audiences:} The linguistic properties can be easily explained to non-experts, however \cliii{metric} selection should not \clii{solely} be performed based on linguistic properties
and  
is advised to be handled by domain experts \revision{for MT and MT metrics}. The methods are especially useful to debug metrics, i.e., for metric developers.\\\textbf{Limitations:} The approach by \citet{kaster-etal-2021-global} could not explain reference-free metrics well, so plausibly requires alternate explanatory variables.  
The search for regressors may be inspired by quality aspect approaches (described earlier this section) where not a global metric score is reported but several scores (such as coherence, fluency, etc.). These could then \clii{compose} a global MT score. Further, some of the ``property metrics'' could be considered black box variables themselves and future work might replace them by more transparent factors. 
One might also explore the collinearity of the different regressors.  
\cliii{For \citet{kaster-etal-2021-global}, faithfulness is likely given to some degree, around the samples that was trained on. Still, it could be advisable to employ faithfulness tests, if possible, to achieve guarantees.}
}

\section{Future Work}
\label{sec:future}

\begin{figure}[htbp]
    \centering
    \resizebox{\textwidth}{!}{\begin{tikzpicture} [set/.style = {draw}]
\node (A)[set, align=left, rectangle,line width=0.7mm] at (0, 6) {\textbf{Counterfactuals}\\``Die Dinge \green{wurden} kompliziert''$\shortrightarrow$Score:1\\``Die Dinge \green{erhalten} kompliziert''$\shortrightarrow$Score:0.5};
\node (B)[set, align=left, rectangle, line width=0.7mm] at (-8, 1.5) {\textbf{Natural Language Explanations}\\In German, the verb "bekommen" means "to get" or\\ "to receive," but in this context, it doesn't convey the\\ intended meaning effectively. Instead, we can use the\\ verb "werden" which means "to become" to better \\capture the sense of change or transformation.};
\node (C)[set, align=left, rectangle, line width=0.7mm] at (6, 3) {\textbf{Decision Rules}\\``became''$\in$ Source \& ``werden'' $\in$ Hypothesis\\$\shortrightarrow$Score$>0.5$};
\node (D)[set, align=left, rectangle, line width=0.7mm] at (-6,-3) {\textbf{Grouped/Hierarchical Feature Importance}\\Importance((``became'',``bekommen'')): 0.1\\Importance((``Things'',``Die Dinge'')): 0.6};
\node (E)[set, align=left, line width=0.7mm] at (5.7,-2.7) {\textbf{Neuron Explanations}\\Neuron 1423: bekommen\\Neuron 5234: werden};
\node (F)[set, align=left, circle,line width=0.7mm] at (0,0) {\textbf{Exemplary Future}\\\textbf{Explanation Types}};
\node (Example)[set, align=left, rectangle, dotted] at (-9, 5.8) {\textbf{Example Data}:\\
\textbf{Ground Truth}: ``Things became complicated''\\
\textbf{Hypothesis}: ``Die Dinge bekamen kompliziert''\\
\textbf{Score}: $0.8$};

\draw[{[sep=6.2pt]}-{Computer Modern Rightarrow[length=3mm, width=3mm]}, line width=0.7mm, shorten <=8pt, shorten >=8pt]
(A) edge (F) (B) edge (F) (C) edge (F) (D) edge (F) (E) edge (F) ;
\end{tikzpicture}}
    \caption{Exemplary future explanation types for MT metrics. \cliii{For each type we provide a hypothetical example. For this we assume the exemplary translation that is shown in the top-left box. Here, the polysemic word "became" is translated with the wrong word sense "to receive" instead of "to get/to develop into". The correct translation is "wurden".}}
\label{fig:futureExTypes}
\end{figure}
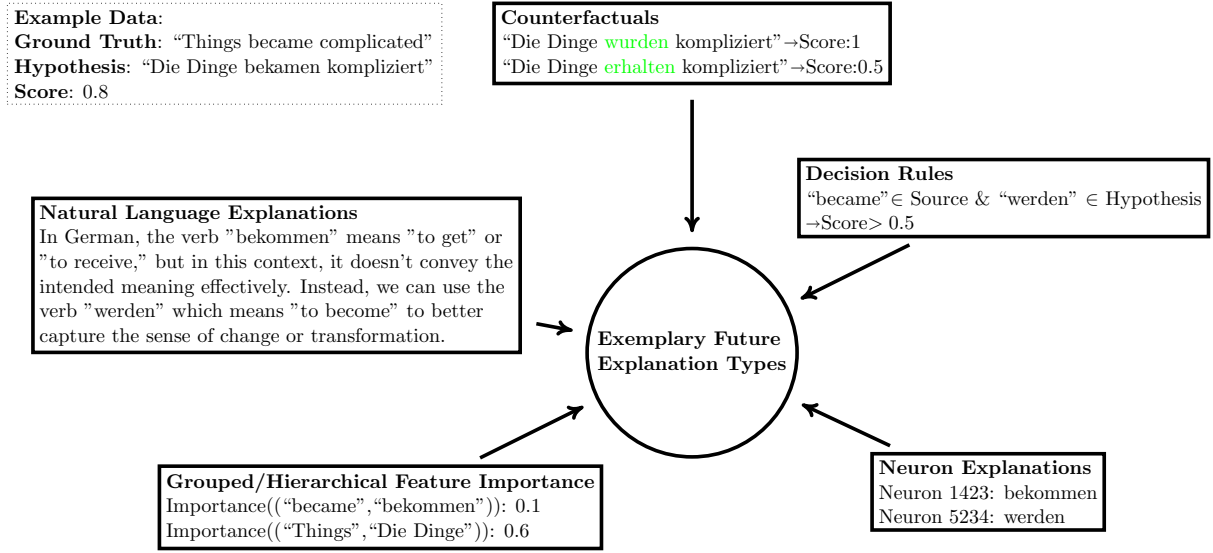

In this section, we describe future research directions for explainable MT metrics. Specifically, we (1) consider directions for explanation types that have been used in the papers described in our taxonomy, (2) describe potential explanation types from other domains, (3) point out needs for future evaluations and concepts and (4) urge for an explanation of domain boundaries.

\noindent\textit{\\Exploring present explanation types} \;\;\;\; Based on our taxonomy (see Figure \ref{fig:taxonomy}), we identify unexplored areas of explainable MT evaluation (empty table cells). 
We find that only \textit{feature importance techniques} are thoroughly explored on all levels of model access. The other methods \clii{mostly explain in black box settings} or are based on metrics that are explainable by design. We suppose that gradients, attention weights, and embeddings have, so far, mostly been applied in the field of feature importance, as there is often already a one-to-one relationship between their values and input features. Also, due to the recent shared tasks, feature importance has likely received the most attention. 

Future work could consider enhancing \cl{\textit{quality aspect explanations}} --- \clii{Row 1 in Figure \ref{fig:taxonomy}} --- by incorporating more model access.
For example, a separate fluency score might be realizable by employing measurements on target side embeddings in unsupervised matching metrics.
Current \cl{model-understanding} techniques \cl{for MT metrics} --- \clii{Row 4-5 in Figure \ref{fig:taxonomy}} --- are \clii{mostly} applied in a black box manner. \cl{For these, using more} model access might also provide further insights. For example, analyses of the contextualized embedding space and how it affects a metric score could explain the metric on a more general level. Many works consider the probing of transformer-based architectures \citep[e.g.,][]{tenney2018what,liu-etal-2019-linguistic, sajjad-etal-2022-analyzing} and it would be interesting to see if their results can inform the metric development.

Also, current perturbation robustness techniques often use manually defined perturbation strategies. It would be interesting to use adversarial attacks instead that automatically infer these strategies. 

\noindent\textit{\\Exploring other explanation types} \;\;\;\; Besides the works in our taxonomy, other fields of NLP and explainable AI in general have proposed further explanation types, e.g., counterfactuals, natural language explanations and rule extraction \citep[e.g.,][]{madsen-etal-2022-post}. We illustrate these options with an example in Figure \ref{fig:futureExTypes}. We already introduced \textit{counterfactuals} in \S\ref{sec:explainability}, stating their role in evaluations of robustness (as adversarial attacks) and fairness. It remains to explore if single sentence explanations also carry explanatory value.
\cl{In the examples for counterfactuals in Figure \ref{fig:futureExTypes}}, the original word ``bekommen'' is switched with the words ``wurden'' (to get) resp.\ ``erhalten'' (to receive) in two counterfactuals.  Based on the scores achieved by these counterfactuals, a user could infer metric weaknesses. One technique to generate counterfactuals is for example described by \citet{wu-etal-2021-polyjuice}.

\textit{Decision rule} explanations describe rules which a model follows when computing its output \cite{ribeiro-2018}. Often, these techniques were designed to explain classification models and are based on \cliii{perturbations} that change the originally predicted class \cite{ribeiro-2018}. For metrics, the scores could be discretized beforehand. An example from other NLP domains are Anchors \cite{ribeiro-2018}. \cliii{In our example (see Figure \ref{fig:futureExTypes}), the rule describes that if ``became'' is present in the source and ``wurden'' is present in the translation, the metric will assign a score higher than 0.5. Obviously in other contexts, such a rule would fail.}

Another type of explanations are \textit{hierarchical feature importance} explanations, which will assign importance scores to groups of input tokens in single \cite{chen-etal-2020-generating-hierarchical} or paired input sentences \cite{chen-2021}. In the given hypothetical example \cl{(see Figure \ref{fig:futureExTypes})}, the groups of \textit{became/bekommen} and \textit{Things/Die Dinge} were formed and get assigned separate importance scores.
A related approach was tested in the SemEval-2015 and -2016 shared tasks \se{on semantic similarity}, where \cliii{the task was to assign labels} that capture the similarity between token groups \cliii{from} two sentences \cite{agirre-2015, agirre-2016}. 

\textit{Natural language explanations} are explanations in free text form \cite[e.g.,][]{wiegreffe-2021}. \cl{This is an explainability type where the generative abilities of recent large language models (LLM), like ChatGPT and GPT4, can be leveraged.} 
The works by \citet{jinlan-etal-2023-gptscore}, \citet{lu2023error}, \citet{kocmi2023large}, \citet{kocmi-federmann-2023-gemba}, \citet{fernandes-etal-2023-devil} and \citet{xu2023instructscore} already explore the use of LLMs as metrics and extract the metric scores from LLM outputs (or generation probabilities) that contain (or could easily be extended to provide) natural language descriptions of why a certain score is assigned. In our taxonomy, we grouped them under fine-grained error explanations, as this is their main goal.\footnote{\cliii{We again note that these techniques will not necessarily describe a metric's internal workings, but might be helpful due to providing plausible explanations, see \S\ref{sec:explainability}.}}  In these works, the explanations are mostly used \se{as} a byproduct\cliii{, while} a thorough evaluation of the explanation quality and further use cases remains \se{for} future work. 

The example in Figure \ref{fig:futureExTypes} was generated with the help of ChatGPT, which shows that such an approach might yield plausible explanations.\footnote{We used ChatGPT in June 2023 with following prompts and asked to rephrase the output once:\\Ground Truth: ``Things got complicate''\\
Hypothesis: ``Die Dinge bekamen kompliziert''\\
Is this translation of the ground truth correct? Why or why not?}
An example from other NLP domains is \citet{rajani-etal-2019-explain}, who generate explanations for commonsense reasoning.
\cliii{In NLG in general, many other approaches for LLM-based metrics have been constructed since the release of ChatGPT. These could potentially be extended to MT metrics and explainability \citep[e.g.,][]{chiang2023large, wang2023chatgpt, ji2023exploring, chen2023exploring, liu2023geval}.}
\revision{As for example \citet{kocmi-federmann-2023-gemba} point out, it it dangerous for academia to rely on proprietary models, especially for evaluation purposes. This includes the following issues: Metrics based on proprietary LLMs (1) could be influenced by the LLM providers, (2) could change or become unavailable at any time, (3) could carry more unexpected biases due to unknown training data and (4) could be contaminated in their evaluation by being trained on common test sets.}

\cliii{Recent works, describe prompting strategies like chain-of-thought \citep{wei2022chain} and tree-of-thought \citep{yao2023tree} where language models show performance improvements if they are giving explanations of their output. While some of the works that are described in our taxonomy already employ such techniques to improve metrics a systematic exploration of the effect on metric scores, or whether these techniques can provide reasonable explanations (that are at least plausible), remains. A first work that considers this systematic exploration is the EVAL4NLP 2023 shared task \citep{leiter2023eval4nlp} where participants should improve metrics scores only by prompting. Their results on a small scale experiment indicate that the participants' generated natural language explanations are not helpful to humans, in most cases.}

\cl{\textit{Neuron Explanations} \se{are} another potential use case \clii{of} LLMs to \clii{separately explain} each Neuron in the underlying networks. \citet{Bills23} recently showed that GPT4 can (to some extent) summarize highlighted neuron activation in textual input of GPT2 to the same extent humans can. This allows to successfully simulate the behavior and effects of some neurons. Interestingly, they also find error related neurons, like a ``post-typo-neuron''. For MT metrics, this would allow to determine the role certain neurons play to determine scores when specific features like ``became is used'' are present in the metric input. At the moment, the success of neuron explanation methods is still limited \citep{Bills23}.}

\cl{\textit{Interactive Explanations:} Due to their context size, instruction fine-tuning and emergent abilities \se{of} recent LLMs allow to interactively ask questions and adapt based on the answers \cite[e.g.,][]{chatgpt}. \citet{jacovi2023diagnosing} state that interactivity is a key point of successfully communicating the explained subject to an explainee. LLM-based explanations (like natural language/neuron explanations ...) might achieve this. \cliii{For example}, a user could ask for more detailed information for a \se{mistranslation} that was noted in a first explanation.}

New approaches could further consider displaying multiple explanation types together, like \citet{xu2023instructscore} who present fine-grained error annotations with natural language explanations.

\noindent\textit{\\Future evaluation and concepts} \;\;\;\; Current work scarcely performs human evaluation of explanations. 
\se{This is} partly \se{because} the focus \se{is} often not on the explanations but other aspects, like the sentence-level scores. One form of human evaluation (for plausibility) is often to use pre-computed human explanations and check the correlation to them \cite[e.g.,][]{zerva-EtAl-2022-WMT}. Current works on MT metrics do, however, not evaluate whether the explanations are actually helpful to explain the concepts they describe to their audience. \clii{This means}, whether the explanations manage to help the explainees to adapt their own theory of mind to be a coherent abstraction of the real problem \cite{jacovi2023diagnosing}. Hence, we encourage future work to address this, e.g., by considering simulatability \cite{hase-2020}. \revision{Human evaluation may also be performed in downstream tasks, by checking performance with and without explanations.} \cliii{Regarding plausibility, it could also be interesting to evaluate whether plausible explanations of incorrect metric scores can actually deceive non-experts or \revision{even translation} experts, as warned by \citet{jacovi-goldberg-2021-aligning} and \citet{Jin2023RethinkingAE} \revision{(for general explainability use cases)}.}

As noted throughout the paper, the degree of faithfulness an explanation has to the metric's internal decision process is an important property \citep{jacovi-2020}, especially for metric debugging and bias detection. Current works on \textit{decision-understanding} of MT metrics, however, mostly dismiss faithfulness. 
Future work might consider the evaluation of metric faithfulness, to enable these use-cases. We note that this evaluation will be complex, due to faithfulness being multi-faceted \citep{jacovi-2020}. For example, lower-granularity scores can faithfully explain sentence-level scores that are aggregated from them, however this does not explain how the actual black box component produced the lower-granularity scores. Further, there are evaluation pitfalls that exist in faithfulness evaluation \citep{ju-etal-2022-logic}, ruling out common methods like area under the perturbation curve scores (AOPC). Therefore, future work should first consider what constitutes faithful explanations in terms of metrics and design evaluation approaches.

Further, we note that current works mostly consider explaining sentence-level scores. Future work, might also consider other metric granularity types more\revision{, such as word-level and document-level metrics. Additionally, future work could explore the influence of language resource availability on the explanations (although it may largely depend on metric quality) and whether some explainable metric setups are not applicable in low-resource settings.}

\noindent\textit{\\Exploring domain boundaries} \;\;\;\;
\cl{Future work might also consider under-explored use cases of explainability for MT metrics. Currently, only one work considers the social biases of MT metrics \cite{sun-etal-2022-bertscore} and none considers the usage of explainability for semi-automatic labeling of training data.}
Besides examining just the metrics, future work could also inspect the relation of explainable MT metrics, especially plausible-only explanations, to explainable MT \citep[e.g.,][]{stahlberg-etal-2018-operation}, as explanations of both will probably be similar.
Lastly, future work could explore providing similar taxonomies as ours that are comprising \emph{the whole field of explainable natural language generation metrics}. 

\section{Related Work} \label{sec:relatedwork}
\cliii{A large number of conceptual works and surveys have been conducted in the area of explainability \citep[e.g.,][]{lipton-2016, biran-2017, adadi-2018, doshi-velez-2017, dosilovic-2018, guidotti-2018, arya-2019, carvalho-2019, gilpin-2019, miller-2019, barredo-arrieta-2020, bodria-2021, linardatos-2021}.}

There are few surveys that touch upon the need for explainable NLG \cliii{(here MT), thereby motivating our work}.
In their survey of NLG metrics, \citet{celikyilmaz-2020} see the need for explainable evaluation metrics to spot system quality issues and to achieve higher trust in the evaluation of NLG systems. They consider that such quality issues might be unintended biases and factual inconsistencies. \citet{sai-2020} instead propose that explainable NLG metrics should focus on providing more information than just a single score (such as fluency or adequacy). \citet{gehrmann2022repairing} also notice the need for interpreting NLG metrics and give a brief overview of selected related approaches. The main contribution of their work is an overview of common NLG metric shortcomings and best practices to prevent them. Our focus on concepts for explainable MT metrics is a different one and our taxonomy of approaches is much more extensive.  

\cliii{General surveys on explainable NLP are related to our work, as they highlight how explainability techniques have been used in NLP and might further guide the development of future approaches for explainable MT metrics \citep[e.g.,][]{Vijayakumar2023InterpretabilityIA,10.1162/tacl_a_00519,Gurrapu2023RationalizationFE,lertvittayakumjorn-2021,Saha2022ExplainabilityOT,Sun2021InterpretingDL,10.1145/3529755,10.1145/3397482.3450728,madsen-etal-2022-post}. 
Besides these works, there are also surveys of robustness \citep[e.g.,][]{10.1145/3593042} and fairness \citep[e.g.,][]{blodgett-etal-2020-language}, which are related, as discussed in \S\ref{sec:explainability}.}

The taxonomy graphic in Figure \ref{fig:taxonomy} \cliii{was adapted from the survey of post-hoc explainability methods in NLP by} \citet{madsen-etal-2022-post}. In specific, we follow their approach to display model access on the x-axis and explanation types on the y-axis. Further, we have customized their TIKZ code. We specifically focus on the explainability of MT metrics; therefore, our taxonomy is completely different content-wise. Our definitions of explainabilty, its goals and audiences are in some parts adapted from the survey of \citet{barredo-arrieta-2020} (see \S\ref{sec:explainability}). Our work is different from previous surveys, as we set our focus on the explainability of MT metrics.

\section{Conclusion} \label{sec:conclusion}
In this work, we discuss audiences, goals and properties for \emph{explainable machine translation metrics}, a nascent field that may help \cliii{further} overcome the dominance of classical low-quality evaluation metrics. We also survey and categorize recent approaches on explainable MT metrics into a taxonomy, highlighting their results and limitations. Currently, two dominant approaches to explainability for MT metrics are (1) feature importance explanations that highlight erroneous words in source and hypothesis to explain sentence-level scores and (2) perturbation robustness approaches that check a metric's robustness to manually devised types of input perturbations. We also identify a current trend towards (3) fine-grained error explanations likely caused by the recent improvements of LLMs like ChatGPT and by the adaptation of MQM annotations in the WMT metrics shared tasks. A \cliii{m}ajor weakness of the current realization of (1) is that error highlights do not consider the correspondence between words in source and target. Further, a weakness of (2) is that perturbation types have to be manually defined, an issue that might be tackled by using more general approaches of adversarial attacks in the future (we discuss this issue in more detail in \sen{an earlier version of this paper; }
\citealp{leiter2022explainable}).
\cliii{Generally, most current evaluations of explainability approaches for MT metrics do not consider the faithfulness of explanations, limiting their use in metric debugging and high-risk scenarios.}
We also present a vision of future approaches to explainable evaluation metrics, which should help fix the problems of the above paradigms and provide guidance to explore unexplored areas. 
Here, we also urge future work to consider the desiderata and implications of faithfulness for explainable MT metrics.  

Our broader vision is that explainability is now a `desirable but optional' feature,
but we argue that in the future it will become 
essential, even compulsory, 
especially for evaluation metrics as a highly sensitive task assessing the quality (and veracity) of translated information content. 
Explainability builds transparency and trust for users, eases bug-fixing and shortens improvement cycles for metric designers and will be required by law/regulations for AI systems to be applied to large-scale, high-stake domains. In this context, we hope our work will 
catalyze efforts on the topic of explainable evaluation metrics for machine translation. 

\acks{Since November 2022 Christoph Leiter is financed by the BMBF project ``Metrics4NLG''. Piyawat Lertvittayakumjorn had been financially supported by Anandamahidol Foundation, Thailand, from 2015-2021. He mainly contributed to this work only until September 2022 while affiliated with Imperial College London (before joining Google as a research scientist after that). Marina Fomicheva mainly contributed to this work until April 2022. 
Wei Zhao was supported by the Klaus Tschira Foundation and Young Marsilius Fellowship, Heidelberg, until December 2023.
Yang Gao mainly contributed to this work before he joined Google Research in December 2021. Steffen Eger is financed by DFG Heisenberg grant EG 375/5–1 and by the BMBF propject ``Metrics4NLG''. }

\appendix

\bibliography{references}

\end{document}